\documentclass[journal]{IEEEtran}
\usepackage[table]{xcolor}
\usepackage{amsmath,amsfonts}
\usepackage{algorithmic}
\usepackage{array}
\usepackage{textcomp}
\usepackage{stfloats}
\usepackage{verbatim}
\usepackage{graphicx}
\usepackage{subfigure}
\usepackage{subcaption}
\usepackage{threeparttable}
\usepackage{multirow}
\usepackage{booktabs}
\definecolor{grn}{rgb}{0,0.7,0}
\newcommand{\std}[1]{{\tiny $\pm$#1}}
\newcommand{\best}[1]{\textbf{\textcolor{red}{#1}}}
\newcommand{\secbest}[1]{\textcolor{blue}{#1}}
\usepackage[colorlinks,linkcolor=grn,anchorcolor=grn,citecolor=grn,urlcolor=black,CJKbookmarks=True]{hyperref}
\def\BibTeX{{\rm B\kern-.05em{\sc i\kern-.025em b}\kern-.08em
    T\kern-.1667em\lower.7ex\hbox{E}\kern-.125emX}}
\usepackage{balance}
\usepackage{amssymb}

\def\model{CAPS}
\def\extendedmodel{FedCAPS}
\begin{document}

\title{Permutation-Invariant Representation Learning for Robust and Privacy-Preserving Feature Selection}

\author{Rui Liu,
        Tao Zhe, 
        Yanjie Fu,~\IEEEmembership{Senior Member,~IEEE}, \\
        Feng Xia,~\IEEEmembership{Senior Member,~IEEE},
        Ted Senator,
        Dongjie Wang
\thanks{Manuscript created October, 2025. \emph{(Corresponding author: Dongjie Wang.)}}
\thanks{Rui Liu, Tao Zhe and Dongjie Wang are with Department of Electrical Engineering and Computer Science, University of Kansas. (e-mail: rayliu@ku.edu, taozhe@ku.edu, wangdongjie@ku.edu)}
\thanks{Yanjie Fu and Ted Senator are with the School of Computing and AI, Arizona State University. (e-mail: yanjie.fu@asu.edu, Ted.Senator@asu.edu)}
\thanks{Feng Xia is with School of Computing Technologies, RMIT University, Melbourne, VIC 3000, Australia (e-mail: f.xia@ieee.org).
}
}

\maketitle

\begin{abstract}
Feature selection eliminates redundancy among features to improve downstream task performance while reducing computational overhead.
Existing methods often struggle to capture intricate feature interactions and adapt across diverse application scenarios.
Recent advances employ generative intelligence to alleviate these drawbacks.
However, these methods remain constrained by permutation sensitivity in embedding and reliance on convexity assumptions in gradient-based search.
To address these limitations, our initial work introduces a novel framework that integrates permutation-invariant embedding with policy-guided search.
Although effective, it still left opportunities to adapt to realistic distributed scenarios.
In practice, data across local clients is highly imbalanced, heterogeneous and constrained by strict privacy regulations, limiting direct sharing.
These challenges highlight the need for a framework that can integrate feature selection knowledge across clients without exposing sensitive information.
In this extended journal version, we advance the framework from two perspectives: 
1) developing a privacy-preserving knowledge fusion strategy to derive a unified representation space without sharing sensitive raw data.
2) incorporating a sample-aware weighting strategy to address distributional imbalance among heterogeneous local clients.
Extensive experiments validate the effectiveness, robustness, and efficiency of our framework. 
The results further demonstrate its strong generalization ability in federated learning scenarios.
The code and data are publicly available \url{https://anonymous.4open.science/r/FedCAPS-08BF}.

\end{abstract}


\section{Introduction}

\IEEEPARstart{F}{eature} selection removes redundant and irrelevant features to improve both predictive performance and computational efficiency in downstream tasks.
Despite the growing dominance of deep learning, feature selection remains indispensable in scenarios characterized by high-dimensional data, the need for interpretability, and limited resource constraints.
For instance, in healthcare, identifying key biomarkers from extensive genetic datasets enhances model transparency and supports informed decision-making.
In finance, extracting relevant features from transactional data facilitates the development of efficient and interpretable risk models.
These examples highlight the importance of feature selection~in~addressing real-world challenges, even in the era of deep learning.

Existing feature selection methods can be divided into three categories:
1) filter methods~\cite{kbest,mrmr} rank features based on specific scoring criteria.
They often select the top-K ranking features as the optimal subset;
2) wrapper methods~\cite{liu2021efficient,gfs} iteratively generate and evaluate feature subsets.
They select the feature subset that performs best on a downstream task as the optimal one;
3) embedded methods~\cite{lasso,sugumaran2007feature} integrate a feature selection regularization term (e.g., $L_1$ regularization) into model training. 
As the model converges, irrelevant features are penalized, and the non-zero coefficients indicate the optimal subset.
While prior work has achieved notable success, it falls to capture complex feature interactions and adapt feature selection knowledge to diverse and~dynamic~scenarios.

Recently, the success of generative AI has inspired researchers to revisit the feature selection problem through a generative lens. By embedding discrete feature selection knowledge into a continuous embedding space, they not only capture complex feature interactions for generating enhanced feature subsets but also make the knowledge more adaptable to diverse scenarios~\cite{gains,ying2024feature,ying2023selfoptimizingfeaturegenerationcategorical,ying2024unsupervisedgenerativefeaturetransformation,azim2024featureinteractionawareautomated,hu2024reinforcementfeaturetransformationpolymer,gao2025gpt}.
However, there are two primary limitations: 
\begin{enumerate}
    \item \textbf{Limitations 1: Permutation bias bring noise into the embedding space and lead to suboptimal performance.}
    Existing methods fail to encode the fact that feature order does not impact model performance within the learned feature subset embedding. 
    This oversight introduces bias into the embedding space, limiting the effectiveness of the search process in discovering optimal feature subsets.

    \item \textbf{Limitation 2: Convexity assumptions of the embedding space hinder effective exploration and search process.}
    Prior literature often assumes that the embedding space is convex, with the expectation that gradient-based search will locate optimal solutions. 
    In practice, this assumption rarely holds, which causes the search process to converge to suboptimal points and produce inferior feature subsets.
    
\end{enumerate}

In the preliminary work~\cite{caps}, we propose \textbf{\model}, a new centralized framework that performs \textbf{\underline{C}}ontinuous optimization for fe\textbf{\underline{A}}ture selection by integrating \textbf{\underline{P}}ermutation-invariant embeddings with a policy-guided \textbf{\underline{S}}earch strategy.
Specifically, given a set of feature selection records, where each record contains the indices of features within a subset and the corresponding model performance, we first design an encoder-decoder framework to capture the underlying patterns in the feature subset indices.
To achieve permutation-invariant embeddings for feature subsets, we leverage a self-attention mechanism that symmetrically computes attention scores across all input feature indices. 
This design guarantees that any permutation of the input features produces identical embeddings, ensuring robust and reliable representations of feature subset indices.
Considering the high computational complexity $O(N^2)$ of pairwise attention calculation, we introduce a set of inducing points to alleviate the computational burden and accelerate processing. 
These inducing points act as intermediate representations, facilitating more efficient attention computations by reducing the need for full pairwise attention. 
This approach achieves a complexity of $O(NM)$, where $M$ denotes the number of inducing points, which is substantially smaller than $N$.
After the encoder-decoder model converges, we deploy a policy-based reinforcement learning (RL) agent to explore the learned embedding space and identify the optimal feature subset.
Based on model performance, we first select the top-K feature subsets as search seeds and input them into the well-trained encoder to obtain their corresponding embeddings.
Next, we employ an RL agent to learn how to optimize these embeddings by maximizing the downstream task performance and minimizing the length of the feature subset.
Throughout this process, the exploratory nature of the RL agent helps overcome the challenges of the non-convex embedding space, enabling the search for improved embeddings.

While the initial framework has demonstrated promising improved outcomes, there remains considerable scope for further improvement. 
In real-world scenarios, data is rarely centralized but instead distributed across multiple local sources, while containing highly sensitive information.
For instance, in the healthcare domain, hospitals store patient records that include personal demographics, medical histories, and treatment outcomes. 
In the financial sector, institutions maintain transaction data and customer profiles that must remain confidential. 
Sharing such raw data across clients introduces serious privacy risks and conflicts with regulatory requirements.
Moreover, data distributions across different local clients are typically heterogeneous and imbalanced. 
For example, hospitals may record different attributes for the same disease depending on their equipment or data collection protocols, while financial institutions may store transaction records with varying sample sizes and feature coverage across branches. 
Directly aggregating such information introduces bias into the global embedding space since clients with large datasets may dominate while smaller clients are underrepresented. 
These inconsistencies across local clients hinder the identification of feature subsets that can generalize across sites. 
Together, these challenges highlight the necessity of developing a framework that aggregates feature knowledge in a privacy-preserving manner.
Conventional federated learning methods (e.g., FedAvg~\cite{fedavg}, FedProx~\cite{fedprox}, FedNTD~\cite{fedntd}, MOON~\cite{moon}) mainly focus on parameter aggregation instead of knowledge fusion.
These methods train local models on private data and iteratively transmit parameter updates to a central server. 
The updates are then aggregated at the server to construct a global model for all participants. 
This paradigm overlooks the fusion of diverse feature selection knowledge from local clients, thereby failing to construct a unified feature embedding space.

To address these challenges, we extend our initial centralized framework and develop {\textbf{\extendedmodel}} (\textbf{Fed}erated \textbf{C}ontinuous optimization for fe\textbf{A}ture selection by integrating \textbf{P}ermutation-invariant embeddings with a policy-guided \textbf{S}earch strategy).
It aims to collaboratively aggregate knowledge from decentralized clients and deriving an optimal feature subset without compromising data privacy.
Specially, we first collect feature selection records for each client individually, where each record contains the feature ID token sequence and its corresponding performance evaluated on the client's local dataset. 
Instead of sharing raw data, each client only transmits its collected feature selection records to the central server, ensuring that sensitive raw data remains local and private. 
The central sever then aggregates the received records into a unified global embedding space using a permutation-invariant encoder-decoder module. 
Within the unified unbiased embedding space, a policy-guided search strategy explores promising regions and identifies optimal feature subsets. 
This RL-based search process integrates a critic that evaluates the performance of agent's action on the clients, reducing communication overhead and improving the efficiency of collaborative optimization.
To further mitigate distributional bias caused by the non-IID and imbalanced client data samples, we adapt a sample-aware weighting strategy during the entire optimization procedure, where clients with larger sample sizes are assigned greater weights. 
Finally, we conduct extensive experiments and comprehensive analyses to evaluate the effectiveness, generalizability and practical utility of our framework against other conventional federated learning baseline models.

\section{Problem Statement}
\noindent The research problem is learning to construct a unified representation space and identify an optimal and generalizable feature subset across decentralized clients without sharing raw data.
Formally, we consider each client holds its own local dataset $\mathcal{D}=\{\mathbf{X},\mathbf{y}\}$, where $\mathbf{X}$ denotes features, and $\mathbf{y}$ represents the corresponding labels. 
Using existing feature selection methods~\cite{marlfs}, we can gather $p$ feature selection records from all local clients, denoted by ${\{(\mathbf{f}_{i},v_{i})\}}^{p}_{i=1}$,  where each record consists of feature indices within the selected subset $\mathbf{f}_{i}$ and their corresponding model performance $v_i$.
Then, we aggregate feature selection knowledge from client records into a global embedding space $\mathcal{E}$ using a permutation-invariant encoder $\omega$ and decoder $\psi$ on the server, optimized by minimizing the reconstruction loss.
Next, a policy-based reinforcement learning (RL) agent is utilized to explore the global embedding space $\mathcal{E}$, aiming to identify the optimal embedding $\mathbf{E}^{*}$. 
This embedding can be used to reconstruct the optimal feature subset $\mathbf{f}^*$ by the decoder $\psi$, which maximizes the weighted average downstream task performance $\mathcal{M}$ across all clients.  
The optimization goal can be formulated as:
\begin{equation}
    \mathbf{f}^{*} = \psi(\mathbf{E}^{*})=\mathrm{argmax}_{\mathbf{E}\in\mathcal{E}}\sum_{c=1}^{C}{\mathcal{W}_{c}\mathcal{M}(X_{c}[\psi(\mathbf{E})])},
\end{equation}
where $\mathcal{W}_{c}$ is the weight of client $c$, $X_{c}$ denotes the local dataset of client $c$.
Finally, we apply feature indices $\mathbf{f}^*$ to local client datasets to generate the optimal feature subsets. 
These feature subsets achieve the maximum weighted global performance.

\section{Methodology}
\subsection{Centralized Model Framework Overview (\model)}
\begin{figure*}[!t]
    \centering
    \includegraphics[width=\linewidth]{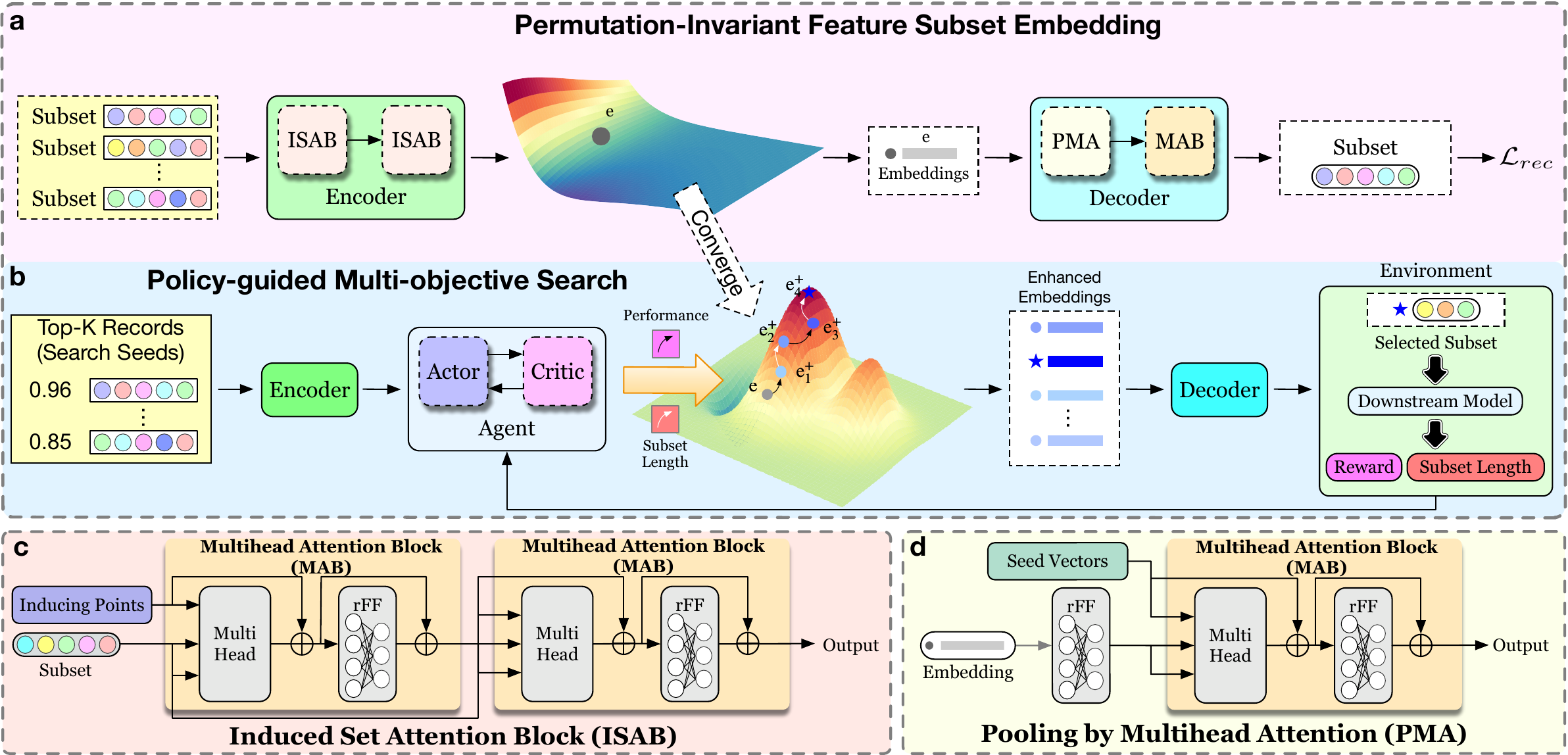}
    \caption{An overview of our centralized model {\model}. First, we develop an encoder-decoder module to learn a permutation-invariant embedding space by optimizing the reconstruction loss (a).
    Then, we explore the learned embedding space through policy-guided RL search, aiming to maximize downstream task performance and minimize feature subset length (b).
    (c) and (d) illustrate the architecture of induced set attention block (ISAB) and pooling by multihead attention (PMA) respectively.}
    \label{framework}
\end{figure*}

\noindent Figure~\ref{framework} illustrates the overview of our initial centralized framework \model, including two components: 
1) permutation-invariant feature subset embedding; 
2) policy-guided multi-objective search.
Specifically, given a large set of feature selection records, each record comprises a selected feature subset and the corresponding model performance.
First, we train the permutation-invariant feature subset embedding module using these records. 
The encoder maps each feature subset to a continuous embedding, while the decoder reconstructs the original feature subset from the learned embeddings.
Second, we select the top-K records based on model performance as search seeds and employ a policy-based RL agent explores the learned embedding space. 
Then, the updated embedding is fed into the well-trained decoder to reconstruct the feature subset. 
The search process is guided by rewards that maximize downstream task performance and minimize the feature subset length.
Finally, the feature subset achieving the highest performance is output as the optimal one. 

\subsubsection{Permutation-Invariant Embedding Learning}
\noindent\textbf{Why permutation-invariant embedding matters.}
Feature subsets are inherently permutation-invariant, as changes in the order of selected features do not impact the downstream task performance. 
Existing works fail to embed this fact into the embedding space, introducing permutation bias. 
To address this, we propose a permutation-invariant encoder-decoder.

\noindent\textbf{Training data collection.}
We first collect $p$ feature selection records as training data, denoted by $\{(\mathbf{f}_{i}, v_{i})\}_{i=1}^{p}$, where $\mathbf{f}_{i}$ is the feature indices of the $i$-th feature subset, and $v_{i}$  denotes the corresponding model performance. 
To achieve this, we use RL-based methods~\cite{marlfs} to automatically explore the feature space. 
In this method, each feature is controlled by an RL agent that decides to select or deselect it. The entire procedure is optimized to maximize downstream task performance. 
For more details, please refer to the referenced paper.

\noindent\textbf{Leveraging a Permutation-Invariant Encoder-Decoder Framework to Embed Feature Subsets.}
In the following contents, we use the indices $\mathbf{f}$ of a feature subset as an example to demonstrate the encoder-decoder calculation process.

\noindent\textbf{Encoder $\omega$ :}
The encoder learns a permutation-invariant mapping function $\omega$ that converts feature indices $\mathbf{f}\in\mathbb{R}^{1\times N}$ into continuous embedding $\mathbf{E}$, denoted by $\omega(\mathbf{f})=\mathbf{E}=[\mathbf{h_1},\mathbf{h_2},\ldots,\mathbf{h_N}]\in\mathbb{R}^{N\times d}$, where $\mathbf{h_n}\in\mathbb{R}^{1\times d}$ is the $t$-th index embedding, $N$ is the number of input features, and $d$ is the hidden size of the embedding. 
Inspired by~\cite{set_tf}, we employ Multihead Attention Block (MAB) to encode pairwise interactions between features in the subset. 
MAB is an adaptation of the Transformer encoder block~\cite{transformer} without positional encoding and dropout, denoted as:
\begin{equation}
    MAB(\mathbf{Q},\mathbf{K}, \mathbf{V}) = LayerNorm(\mathbf{H}+rFF(\mathbf{H})),
\end{equation}
\begin{equation}
    \mathbf{H} = LayerNorm(\mathbf{Q}+Multihead(\mathbf{Q},\mathbf{K},\mathbf{V};\mathbf{W})),
\end{equation}  
where $rFF$ denotes row-wise feedforward layer, $\mathbf{H}$ is an intermediate representation, $LayerNorm$ is layer normalization~\cite{lei2016layer}, $Multihead$ denotes the multihead attention mechanism, and $\mathbf{Q}$, $\mathbf{K}$, and $\mathbf{V}$ denote the query, key and value respectively. 
We set $\mathbf{Q}=\mathbf{K}=\mathbf{V}=\mathbf{f}$ to preserve original information and extract pairwise relationship among all features.
This design focuses on feature relationships rather than their order, thereby enhancing the robustness of the learned embeddings. 
However, the quadratic complexity $O(N^2)$ of pairwise attention becomes prohibitive as $N$ grows.
To mitigate the quadratic cost, we employ a set of inducing points as intermediate representations, encoding global information about the input features. 
Each inducing point serves as a representative anchor in the embedding space, capturing distinct patterns from feature subsets and enabling efficient global aggregation.
This strategy reduces the complexity to $O(NM)$, where $M\ll N$.
Formally, given feature indices $\mathbf{f}\in\mathbb{R}^{1\times N}$ and $M$ inducing points $\mathbf{I}$, the ISAB is defined as
\begin{equation}
    ISAB_{M}(\mathbf{f}) = MAB(\mathbf{f},\mathbf{H},\mathbf{H})\in\mathbb{R}^{N\times d},
\end{equation}
\begin{equation}
    \mathbf{H} = MAB(\mathbf{I},\mathbf{f},\mathbf{f})\in\mathbb{R}^{M\times d},
\end{equation}
where the inducing points $\mathbf{I}\in\mathbb{R}^{M\times d}$ are learnable parameters, and $d$ is the hidden size. 
The ISAB comprises of two MAB layers. 
The first layer uses inducing points $\mathbf{I}$ as queries and feature indices $\mathbf{f}$ as $\mathbf{K}$ and $\mathbf{V}$ to produce a low-dimensional representation $\mathbf{H}$; 
The second layer uses $\mathbf{H}$ as keys and values to inject this global information back into each feature.
To capture high-ordered feature interactions, we stack two ISAB layers to construct our encoder:
\begin{equation}
    \omega(\mathbf{f})=ISAB_{M}(ISAB_{M}(\mathbf{f}))=\mathbf{E}
\end{equation}

\noindent\textbf{Decoder $\psi$ :}
The decoder $\psi$ reconstructs the feature indices $\mathbf{f}$ from the continuous embedding $\mathbf{E}$ in the learned space $\mathcal{E}$, denoted by $\psi(\mathbf{E})=\mathbf{f}$. 
To aggregate information efficiently, we adopt Pooling by Multihead Attention (PMA), which integrates MAB with $K$ learnable seed vectors $\mathbf{S}\in\mathbb{R}^{K\times d}$. 
These seed vectors serve as prototype queries that selectively attend to different aspects of the learned embedding, thereby pooling complementary information. 
Formally, given $\mathbf{E}\in\mathbb{R}^{N\times d}$, a PMA module with $K$ seed vectors is defined as follows:
\begin{equation}
    PMA_{K}(\mathbf{E}) = MAB(\mathbf{S},rFF(\mathbf{E}),rFF(\mathbf{E}))\in\mathbb{R}^{K\times d},
\end{equation}
where $rFF$ denotes row-wise feedforward layer, and $\mathbf{S}$ is jointly learned with the rest of the network. 
Similar to the inducing points $\mathbf{I}$, the seed vectors serve as queries to pool information from $\mathbf{E}$. 
To further capture the interactions among the $K$ pooled outputs, we employ MAB and $rFF$ afterwards:
\begin{equation}
    \mathbf{\Tilde{f}} = rFF(MAB(PMA_{K}(\mathbf{E})))\in\mathbb{R}^{1\times N},
\end{equation}
where $\mathbf{\Tilde{f}}$ is the reconstructed feature indices sequence.

\noindent\textbf{Optimization}. 
The encoder and decoder are trained by minimizing the negative log-likelihood loss of $\mathbf{f}$ given $\mathbf{E}$:
\begin{equation}
    \mathcal{L}_{rec} = - logP_{\psi}(\mathbf{f}|\mathbf{E}) = - \sum^{N}_{n = 1}logP_{\psi}(\mathbf{f}_{n}|\mathbf{h}_{n}).
    \label{reconstruction_loss}
\end{equation}

\subsubsection{Policy-Guided Multi-Objective Search}
\noindent\textbf{Why selecting search seeds matters.} 
As the model converges, we conduct a policy-guided RL search to identify the optimal feature subset embedding. 
Following initialization techniques in deep learning, we start the search from advantageous points, referred to as search seeds, to improve efficiency and performance. 
Specifically, we rank records by model performance and select the top-$K$ as seeds to guide the search in the embedding space.

\noindent\textbf{Policy-guided multi-objective search}. 
After constructing the embedding space, we explore it to identify optimal feature subset embeddings. 
However, this space is generally non-convex due to complex feature interactions.
In addition, searching for an optimal subset is inherently multi-objective, involving trade-offs between maximizing downstream task performance and minimizing subset length.
To address these challenges, we employ Proximal Policy Optimization (PPO)~\cite{schulman2017proximalpolicyoptimizationalgorithms}. 
PPO stabilizes training by clipping updates within trust regions, effectively balancing exploration and exploitation in high-dimensional spaces without relying on convexity assumptions.
The PPO agent takes learned embedding $\mathbf{E}$ as input and outputs an enhanced one $\mathbf{E^{+}}$. 
The decoder $\psi$ then reconstructs $\mathbf{E^{+}}$ into a feature subset $\mathbf{f^{+}}$, which is the agent’s state $\mathbf{s}$. 
The downstream task performance of $\mathbf{f^{+}}$ is used as the reward $\mathbf{R}$. 
Thus, our RL framework consists of agent, state, and reward.

\noindent\underline{$Agent.$} PPO agent $\mathcal{P}$ manipulates the embedding $\mathbf{E}$ to generate an enhanced embedding $\mathbf{E}^{+}$, expressed as:
$
\mathbf{E}^{+} = \mathcal{P}(\mathbf{E})
$.

\noindent\underline{$State.$} State $\mathbf{s}$ describes the current selected feature indices. 
Specifically, we decode enhanced embedding $\mathbf{E}^{+}$ by the well-trained decoder $\psi$ to obtain the state representation: 
$
\mathbf{s} 
    = Rep(X[\psi(\mathbf{E}^{+})]) 
$, where $Rep$ is one-hot encoding approach.

\noindent\underline{Reward.} Reward $\mathbf{R}$ evaluates the searched subset, defined as:
$
    \mathbf{R} = \lambda(\mathcal{M}(X[\mathbf{f}^{+}])-\mathcal{M}(X[\mathbf{f}])) + (1-\lambda)\mathcal{N}[\mathbf{f}^{+}],
    \label{eq:reward}
$
where $\mathcal{M}$ is the downstream ML task, $\mathbf{f}$ is the original feature indices, $\mathbf{f}^{+}$ is the searched ones, $\mathcal{N}[\mathbf{f}^{+}]$ denotes the length of $\mathbf{f}^{+}$, and $\lambda$ is a trade-off hyperparameter that balances improvement in model performance and the length of the selected feature subset.

\noindent\underline{Solving the Optimization Problem.} 
We train the PPO agent with multiple objectives, aiming to maximize downstream performance while minimizing feature subset size.
To this end, we adopt an actor–critic framework with separate actor and critic networks. 
The critic model is optimized by minimizing the error between estimated and cumulative discounted rewards:
\begin{equation}
    \mathcal{L}_{critic} = \frac{1}{T}\sum^{T}_{t=1}(V(\mathbf{s}_t)-G_t)^{2} 
    ,
\end{equation}
where $T$ is the length of the trajectory, $\mathbf{s}_t$ is the state, $V(\mathbf{s}_t)$ is the estimated output of critic, and $G_t$ is the discounted cumulative reward. 
To train the actor network, we maximize a clipped surrogate objective while constraining update steps:
\begin{equation}
    \mathcal{L}_{actor} = \hat{\mathbb{E}}_{t}\left[ min(r_t(\theta)\hat{A}_t, clip(r_t(\theta), 1-\epsilon,1+\epsilon)\hat{A}_t)\right],
\end{equation}
where $\epsilon$ is a hyperparameter, $r_t(\theta)$ is the probability ratio of new and old policy, $\mathbf{a}_t$ is the actor output, $\hat{A}_t$ is the estimated advantage function, and $clip(\cdot)$ prevents $r_t(\theta)$ from leaving $[1-\epsilon,1+\epsilon]$.  
By taking the minimum of clipped and unclipped objective, policy updates remain within a trust region, ensuring stability. 
Once the actor converges, we obtain the optimal policy $\pi^*$ which guides embeddings $\mathbf{E}$ toward directions that improve downstream performance in $\mathcal{E}$.
The well-trained decoder $\psi$ then reconstructs the $K$ enhanced feature subset indices $[\mathbf{f}^{+}_{1},\mathbf{f}^{+}_{2},...,\mathbf{f}^{+}_{K}]$ from the enhanced embedding $[\mathbf{E}^{+}_{1},\mathbf{E}^{+}_{2},...,\mathbf{E}^{+}_{K}]$. 
Based on these indices, we derive feature subsets from the original feature set and evaluate them on the downstream task. 
Finally, we select the feature subset indices that achieves the highest performance as the optimal one $\mathbf{f}^{*}$.

\subsection{Federated Model Framework Overview (\extendedmodel)}
\begin{figure}[!t]
    \centering
    \includegraphics[width=\linewidth]{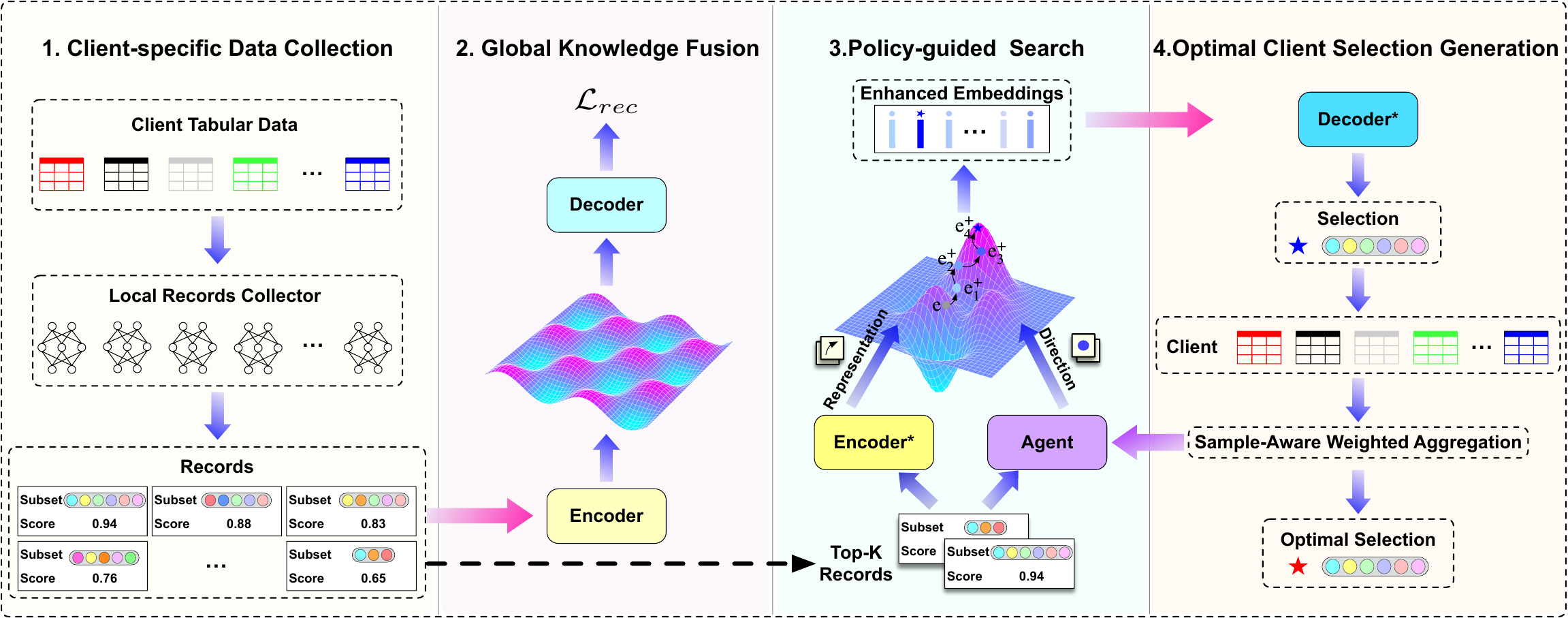}
    \vspace{-0.35cm}
    \caption{An overview of our federated model {\extendedmodel}.}
    \vspace{-0.35cm}
    \label{federated_framework}
\end{figure}

To extend our centralized framework to the federated setting, we develop \extendedmodel, a federated feature selection framework designed to aggregate knowledge across decentralized clients in a privacy-preserving manner. 
Figure~\ref{federated_framework} illustrate an overview of the proposed federated framework \extendedmodel.  
The overall workflow of {\extendedmodel} consists of three main stages. 
First, each client independently collects feature selection records, where each record includes a sequence of feature indices representing the selected subset and its corresponding performance on the local dataset. 
These records are transmitted to the central server as knowledge representations, rather than raw data. 
By transmitting only feature indices and their associated performance, clients contribute client-specific feature selection knowledge. 
In this way, sensitive raw data remains protected and compliance with privacy and regulatory requirements is maintained. 
Second, the central server employs a permutation-invariant encoder–decoder module to integrate the collected records into a unified embedding space.
This embedding space serves as a global representation that aggregates diverse feature selection knowledge across clients while remaining insensitive to the order of feature indices. 
In this way, diverse feature selection knowledge contributed by heterogeneous clients can be fused into a single representation space that facilitates global optimization.
Third, a policy-guided reinforcement learning (RL) agent explores the unified embedding space to identify globally optimal feature subsets.
To initialize the search effectively, we first rank all available records based on their downstream task performance and select the top-$K$ as search seeds.
The agent then iteratively optimizes embeddings in the learned space, guided by multi-objective rewards that encourage both high weighted global performance and compact subsets.
The enhanced embeddings are decoded into candidate feature subsets, which are evaluated across all clients under a sample-aware weighting scheme.
This weighted evaluation strategy mitigates distributional bias and ensures that the final selected subset $\mathbf{f}^*$ achieves the highest weighted global performance.
Through this design, {\extendedmodel} establishes an effective mechanism for collaborative feature selection in federated settings, enabling global optimization with privacy preservation and reduced communication overhead.

\subsubsection{Privacy-preserving Global Knowledge Aggregation}
\noindent\textbf{Why privacy-preserving knowledge aggregation matters.}
In federated feature selection, directly sharing raw data across clients is infeasible due to strict privacy requirements and regulatory constraints.  
For instance, medical institutions cannot expose patient-level records, while financial organizations must protect sensitive customer information.  
These restrictions make data centralization impractical, requiring alternative approaches that preserve privacy while still enabling collaboration. 
Conventional federated learning (FL) methods primarily rely on parameter aggregation to build a global model. 
These approaches train local models for each client on private datasets and on periodically transmit model parameters such as weights or gradients to a central server. 
The server then averages these parameters to update a global model, which is redistributed to clients for further training. 
Despite its success in distributed classification and prediction, this paradigm falls short when applied to federated feature selection. 
Parameter aggregation captures only weight updates from local models, overlooking the diverse feature knowledge within local datasets. 
Thus, the global model fails to reflect which feature subsets are consistently informative across heterogeneous clients.
To address this limitation, we reformulate the aggregation process from a knowledge-centric perspective. 

\noindent\textbf{Privacy-Preserving Global Knowledge Aggregation.}
{\extendedmodel} builds upon the same architectural backbone as the centralized \model, consisting of a permutation-invariant encoder–decoder module and a policy-guided reinforcement learning agent.
In contrast to conventional FL settings, both components operate exclusively on the server side. 
The encoder–decoder module is trained directly on feature selection records collected from local clients. 
The encoder embeds the feature indices sequences into continuous embeddings, aggregating diverse feature knowledge across clients. 
The decoder reconstructs the original feature subsets from the embeddings. 
The architecture of {\extendedmodel} remains identical to the centralized framework \model, but it operates on client-contributed records rather than centralized data. 
The encoder–decoder module is optimized using the same negative log-likelihood loss defined in Eq.~\eqref{reconstruction_loss}. 
In addition to ensuring permutation invariance, this design aligns heterogeneous client knowledge within a unified global space, providing the foundation for subsequent policy-guided search. 
To explore this unified embedding space, we adopt the same policy-guided RL strategy as in the centralized framework. 
Specifically, we select the top-$K$ records as search seeds and employ a PPO agent, consisting of an actor that outputs exploration directions and a critic that evaluates actor's actions. 
Unlike the centralized setting, the reward is derived from multiple clients rather than a single-site metric. 
To achieve this, we introduce a sample-aware weighted global performance aggregated over all clients. 
To reduce communication overhead, the critic is trained to approximate this weighted reward using sparse client feedback. 
Thus, most policy updates rely on the outputs of critic network, while periodic client-side evaluations recalibrate the target baseline.

\begin{table*}[!t]
\centering
\vspace{-0.2cm}
\caption{Overall performance comparison of \textbf{centralized framework {\model}}. Best results are in \best{red}, second-best in \secbest{blue}.}
\vspace{-0.2cm}
\label{table_overall_perf}
\resizebox{\linewidth}{!}{
\begin{tabular}{cccccccccccccccccc}
\hline
Dataset          & Task & \#Samples & \#Features & Original & K-Best  & mRMR & LASSO & RFE  & LASSONet &GFS &SARLFS  & MARLFS &RRA & MCDM & MEL & GAINS &\model        \\  \hline
SpectF & C   & 267     & 44 &  75.96 &82.36&  81.41&  79.16&  78.21 &  79.16& 75.01 &75.96&  75.96& 81.23 & 79.16 & 79.69& \secbest{87.38}& \best{89.71} \\  
SVMGuide3 & C   & 1243    & 21&  77.81 &77.02&  77.42&  74.77&  78.90&  76.51& \secbest{83.70} &  78.92&  82.26  & 77.63 & 77.26 & 73.95 & 83.68& \best{84.22} \\  
German Credit      & C   & 1001    & 24 & 64.88 &  68.07&  66.07&  67.58 &  65.42&   66.27& 65.51 & 65.02&  68.17& 66.90 & 68.97& 64.61  &   \secbest{75.34}& \best{77.15} \\ 
Credit Default     & C   & 30000   & 25  & 80.19 &  79.82&  80.38&  77.94&  80.22&   80.44& 80.49&  80.48&  80.47&  75.01& 74.33& 74.89 &  \secbest{80.61}& \best{80.91} \\ 
SpamBase          & C   & 4601    & 57 & 92.68 & 91.34&  92.02&  91.69&  91.80&   89.63& 91.73 &90.94&  91.38&  90.51& 90.83& 90.47 &  \secbest{92.93}& \best{93.13} \\  
Megawatt1  & C & 253 & 38 &  81.60 &80.08&  80.08&  83.78 &  80.08 &  82.83& 78.60 & 82.75&  78.71&  86.05& 82.83& 78.18 &   \secbest{90.42}& \best{91.83} \\
Ionosphere        & C   & 351     & 34 & 92.85 &  92.74&  89.93&  94.14&  95.69&   94.14& 92.77& 89.82&  88.51&   92.72& 92.74& 90.16 &   \secbest{97.10}& \best{98.55} \\  
\hline
Mice-Protein  & MC & 1080 & 77 & 74.99 & 76.37&  78.71&  78.71&  77.29&  75.49& 75.00& 74.53&  75.01& 78.70& 78.27& 76.10 &    \secbest{79.16}& \best{83.32} \\
Coil-20   & MC & 1440 & 400 &  96.53 & 97.58&  96.53&  94.81&  \secbest{97.92} & 94.8& 92.72&   94.79&  95.14& 94.46& 93.42& 95.32 &   97.22& \best{98.27} \\
MNIST fashion   & MC & 10000 & 784 & 80.15 & 79.45&  80.40&  79.50&  80.70&  79.45& 80.15& 80.10&  79.90& 79.95& 79.75& 79.98 &    \secbest{81.00}& \best{81.15} \\
UrbanSound& MC & 5702 & 16000 & 29.71&	28.74&	28.92&	27.52&	29.89&	29.18&	28.57&	30.24&	29.18&	30.50&	29.10& 24.92 &  	\secbest{30.67}&	\best{32.52} \\
\hline
Openml\_589          & R   & 1000    & 25 & 50.95 & 55.30&  55.28&  59.74&  55.03&  56.32& 59.74& 53.43&  53.51& 54.54& 55.43& 53.99 &     \secbest{59.76}& \best{59.77}  \\  
Openml\_616           & R   & 500     & 50 & 15.63 & 24.86&  24.27&  29.52&  23.90&  22.98& 44.44& 25.35&  25.68& 24.67& 24.28& 42.26 &    \secbest{47.39}& \best{48.52} \\
IQ-Dataset & R & 20000 & 38 & 98.54&	\secbest{98.61}&	\secbest{98.61}&	98.20&	98.58&	98.26&	90.95&	98.26&	98.27&	79.20&	98.54& 98.59 &  	98.59& \best{98.68}\\
\hline
\end{tabular}}
\begin{tablenotes}
    \item * We evaluate classification (C), multi-classification (MC), and regression tasks in terms of F1-Score, Micro-F1, and 1-RAE respectively. 
    \item * The higher the value is, the better the model performance is.
\end{tablenotes}
\vspace{-0.5cm}
\end{table*}

\subsubsection{Sample-Aware Weighted Aggregation}
\noindent\textbf{Why sample-aware weighted aggregation matters.}
In federated feature selection, datasets distributed across clients are typically heterogeneous, especially in sample size and statistical stability. 
According to the law of large numbers, clients with larger datasets provide more stable and representative performance estimates.
In contrast, smaller datasets are more likely to yield noisy or biased results. 
When all client records are aggregated equally, the inherent variations in dataset size and stability are overlooked. 
As a result, the global embedding space is biased by unreliable information, leading to lower-quality feature subsets. 
To address this issue, we adopt a sample-aware weighting strategy that assigns larger weights to clients with more samples, thereby reducing distributional bias and improving the representativeness of aggregated knowledge.

\vspace{-0.15cm}
\noindent\textbf{Sample-Aware Weighted Aggregation.} 
To implement the proposed strategy, we design a sample-aware weighting aggregation mechanism that adjusts the contribution of each client based on its dataset size. 
This approach enhances statistical stability and provides a more reliable foundation for subsequent policy optimization. 
During the client-side evaluation phase of PPO, the server employs the well-trained decoders to reconstruct feature indices sequences $[\mathbf{f}^{+}_{1},\mathbf{f}^{+}_{2},\ldots,\mathbf{f}^{+}_{K}]$ from the actor’s enhanced embeddings $[\mathbf{E}^{+}_{1},\mathbf{E}^{+}_{2},\ldots,\mathbf{E}_{K}]$. 
These reconstructed sequences represent candidate feature subsets explored by the PPO agent.
The server then broadcasts the sequences to all participating clients $[c_{1}, c_{2}, \ldots, c_{C}]$, where $C$ denotes the number of clients in the federation.
Each client $c$ applies these sequences to its local dataset to obtain the corresponding downstream task performance scores. 
For a given sequence $\mathbf{f}^{+}_{i}$, this procedure yields a vector of client-specific evaluations $[v_{1}, v_{2}, \ldots, v_{C}]$, where $v_{c}$ denotes the performance measured on $c$-th local client dataset. 
These evaluations capture how consistently a feature subset generalizes across heterogeneous client distributions. 
They serve as a multi-faceted signal that cannot be obtained from any single client in isolation. 
To integrate these evaluations into a unified reward feedback, we adopt a sample-aware weighting scheme. 
Specifically, each client $c$ is assigned a weight proportional to its dataset size: 
$\mathcal{W}_{c} = \frac{\left|\mathcal{D}_{c}\right|}{\sum^{C}_{j = 1}\left|\mathcal{D}_{j}\right|}$, where $\left|\mathcal{D}_{c}\right|$ represents the number of samples held by local client $c$. 
The global performance $\hat{v}_{\textbf{f}_i}$ of feature indices sequence $\textbf{f}_i$ is then computed as: 
$\hat{v}_{\textbf{f}_i} = \sum_{c=1}^{C}\mathcal{W}_{c} \cdot v_{c}$. 
Finally, the sequence that achieves the highest global performance $\hat{v}$ is considered as the optimal one $\textbf{f}^{*}$. 
This design balances contributions across heterogeneous clients by emphasizing reliable evaluations from larger datasets and mitigating noise from smaller ones. 
The resulting aggregated reward not only enhances the generalizability of the selected feature subsets but also establishes a robust foundation for global optimization.
Moreover, critic calibration with ground-truth client feedback ensures stable policy updates and facilitates efficient convergence.
In this way, the proposed weighted aggregation strategy enhances both the effectiveness and reliability of exploration in the unified embedding space.

\begin{table*}[!t]
\centering
\caption{Overall performance comparison of \textbf{federated framework {\extendedmodel}}. Best results are in \best{red}, second-best in \secbest{blue}.}
\label{table_fed_methods}
\resizebox{\linewidth}{!}{
\begin{tabular}{lcc@{\hskip 12pt}cc@{\hskip 12pt}cc@{\hskip 12pt}cc@{\hskip 12pt}cc}
\hline
\multirow{2}{*}{Dataset} 
& \multicolumn{2}{c}{FedAvg} 
& \multicolumn{2}{c}{FedNTD} 
& \multicolumn{2}{c}{FedProx} 
& \multicolumn{2}{c}{MOON} 
& \multicolumn{2}{c}{FedCAPS} \\ 
\cmidrule(lr){2-3} \cmidrule(lr){4-5} \cmidrule(lr){6-7} \cmidrule(lr){8-9} \cmidrule(lr){10-11}
 & Local &  Global & Local &  Global & Local &  Global & Local & Global & Local & Global \\ \hline
SpectF        & \secbest{75.75\std{9.00}} & 79.36 & 75.59\std{5.05} & 78.21 & 74.60\std{10.75} & 75.96 & 74.48\std{3.81} & \secbest{81.23} & \best{76.24}\std{10.27} & \best{87.63} \\
SVMGuide3     & \best{80.31}\std{1.12} & \secbest{83.18} & 79.00\std{4.21}    & 82.77 & 78.26\std{3.93}  & 82.64 & 79.00\std{2.61}    & 79.84 &\secbest{79.10}\std{1.13}  & \best{83.33} \\
German Credit & 79.21\std{7.39} & 70.63 & \best{79.45}\std{1.52} & 70.43 & 79.24\std{3.05}  & 68.96 & 79.04\std{5.44} & \secbest{72.06} & \secbest{79.32}\std{4.52}  & \best{76.65} \\
Credit Default& \secbest{79.92}\std{0.79} & 80.48 & 79.83\std{0.80} & 80.38 & 79.83\std{1.12}  & 80.26 & 79.84\std{0.74} & \best{80.68} & \best{79.96}\std{0.82}  & \secbest{80.57} \\
SpamBase      & 93.62\std{0.19} & 92.04 & \secbest{93.97}\std{1.23} & 91.93 & 93.54\std{1.93}  & 92.15 & 93.64\std{1.37} & \secbest{92.49} & \best{93.98}\std{1.11}  & \best{93.13} \\
Megawatt1     & \secbest{79.26}\std{8.64}& 90.29 & 73.81\std{18.75}& \secbest{90.64} & 74.56\std{13.17} & 87.89 & 77.31\std{5.08}& 82.59 & \best{88.10}\std{3.37}  & \best{91.83} \\
Ionosphere    & \best{92.83}\std{1.98} & \secbest{97.20} & 90.44\std{1.40}& 97.10 & 91.37\std{3.42}  & 95.69 & 92.60\std{4.27} & 94.14 & \secbest{92.76}\std{0.65} & \best{98.55} \\ \hline
Mice-Protein  & \best{87.05}\std{2.83} & \secbest{82.42} & 85.27\std{1.30} & 80.56 & 86.19\std{3.91} & 76.85 & 85.78\std{3.73} & 81.93 & \secbest{86.25}\std{3.26} & \best{83.32} \\
Coil-20       & 83.35\std{1.44} & \secbest{96.54} & \best{84.00}\std{2.42}    & 95.50 & 83.30\std{3.07} & 95.84 & 82.98\std{1.33} & 95.83 & \secbest{83.94}\std{0.92} & \best{97.58} \\
MNIST fashion & \secbest{78.30}\std{0.27} & \secbest{80.55} & 77.85\std{0.92} & 80.10 & 78.15\std{0.18} & 80.15 & 77.90\std{0.93} & 80.20 & \best{79.65}\std{0.23} & \best{81.15} \\
UrbanSound    & \secbest{32.49}\std{4.36} & \secbest{30.24} & 32.41\std{4.24} & 29.18 & 31.79\std{0.68} & 28.49 & 32.05\std{4.46} & 27.69 & \best{32.78}\std{1.66} & \best{30.85} \\ \hline
Openml\_589   & \secbest{36.43}\std{13.33}& 54.67 & 33.46\std{10.90}& 55.93 & 36.31\std{9.87} & \secbest{56.86} & 32.57\std{12.95}& 52.15 & \best{36.98}\std{11.18}& \best{59.90} \\
Openml\_616   & \secbest{10.60}\std{5.84} & \secbest{34.07} & 7.54\std{5.84}  & 28.17 & 7.24\std{4.58}   & 27.33 & 9.00\std{3.58}  & 31.14 & \best{15.30}\std{2.11} & \best{48.52} \\
IQ-Dataset    & 98.35\std{0.06} & \secbest{98.60} & \secbest{98.36}\std{0.06}& 98.58 & \secbest{98.36}\std{0.06}  & 98.57 & 98.35\std{0.07} & 98.58 & \best{98.38}\std{0.03} & \best{98.68} \\
\hline
\end{tabular}}
\begin{tablenotes}
    \item * We evaluate classification (C), multi-classification (MC), and regression tasks in terms of F1-Score, Micro-F1, and 1-RAE respectively. 
    \item * The standard deviation is computed based on the results of 3 local clients.
    \item * The higher the value is, the better the model performance is.
\end{tablenotes}
\end{table*}

\vspace{-0.1cm}
\section{Experiments}

\begin{figure*}[!h]
\subfigure[SVMGuide3 (\model)]{
\includegraphics[width=0.23\linewidth, trim={0 0 1.8cm 0}]{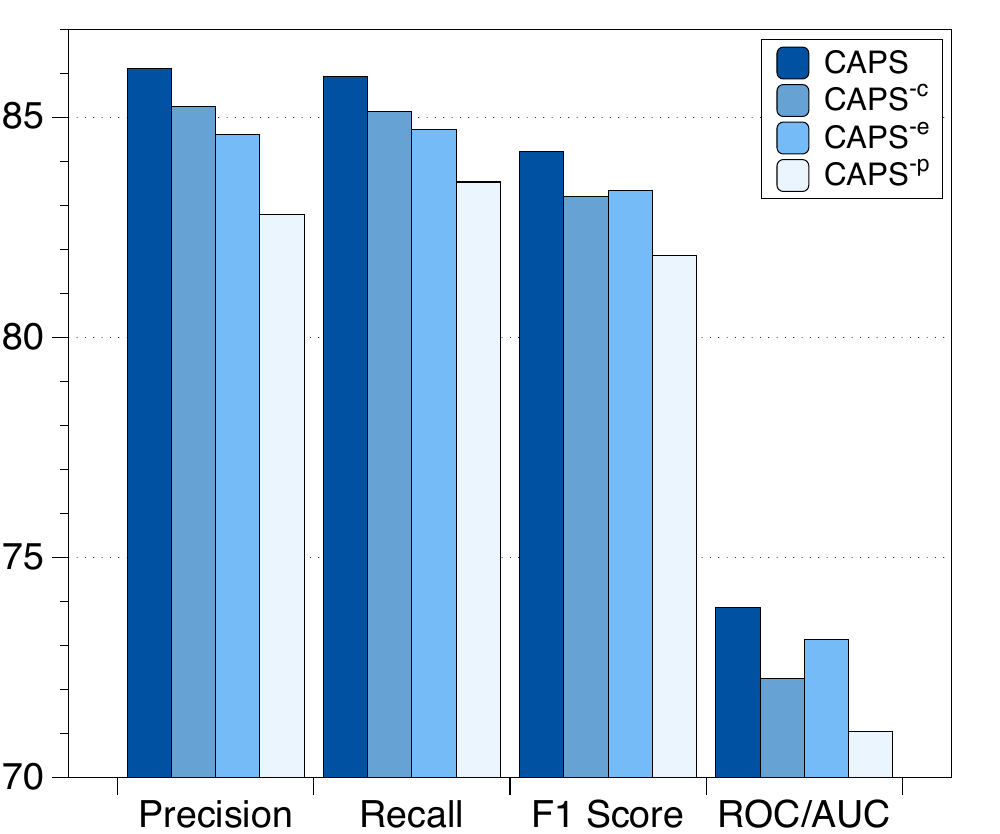}
}
\hspace{-1mm}
\subfigure[German Credit (\model)]{
\includegraphics[width=0.23\linewidth, trim={0 0 1.8cm 0}]{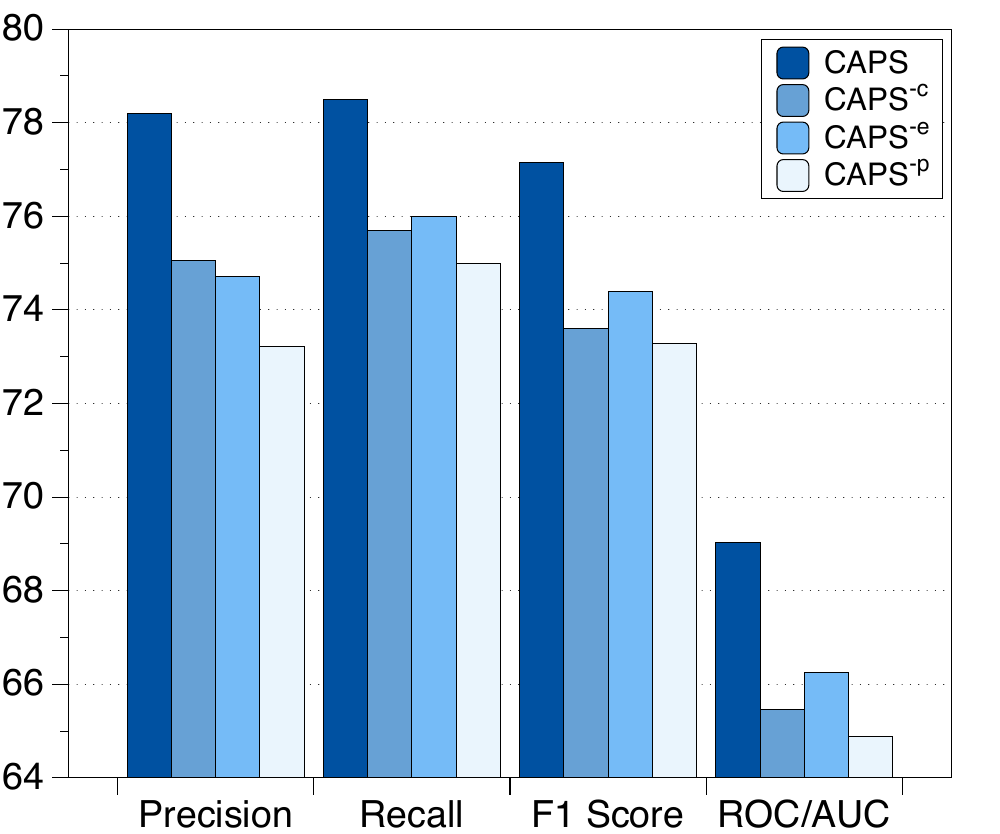}
}
\hspace{-1mm}
\subfigure[Mice Protein (\extendedmodel)]{ 
\includegraphics[width=0.23\linewidth, trim={0 0 1.8cm 0}]{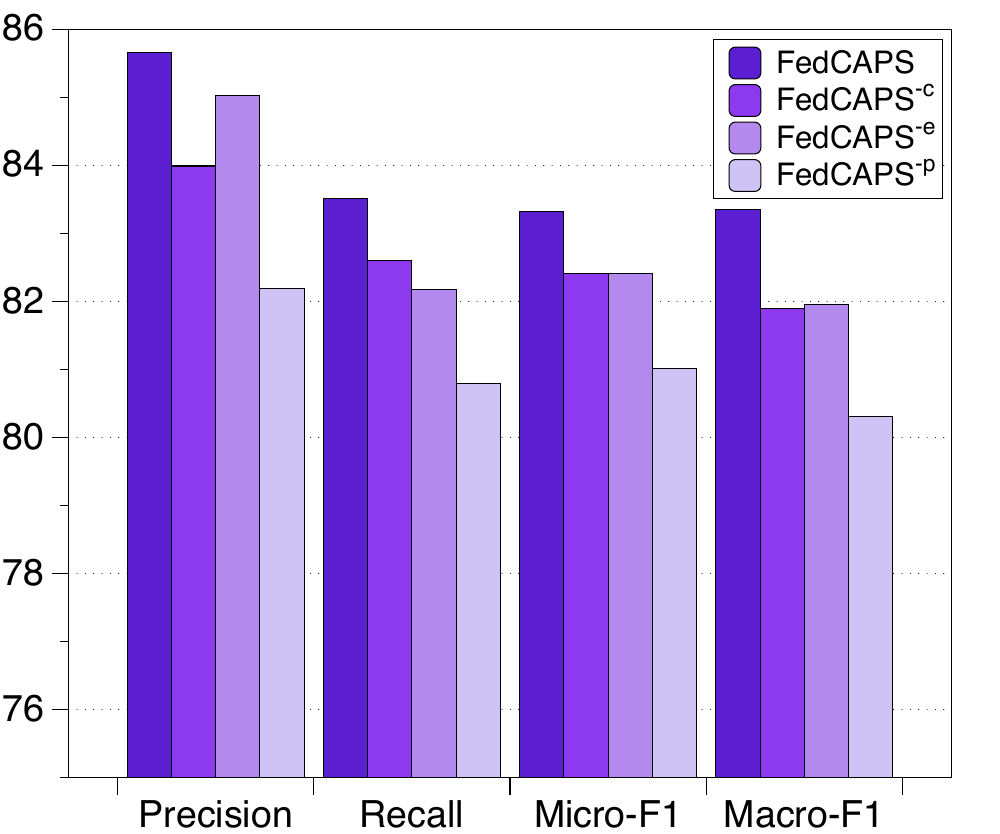}
}
\hspace{-1mm}
\subfigure[Openml\_616 (\extendedmodel)]{ 
\includegraphics[width=0.23\linewidth, trim={0 0 1.8cm 0}]{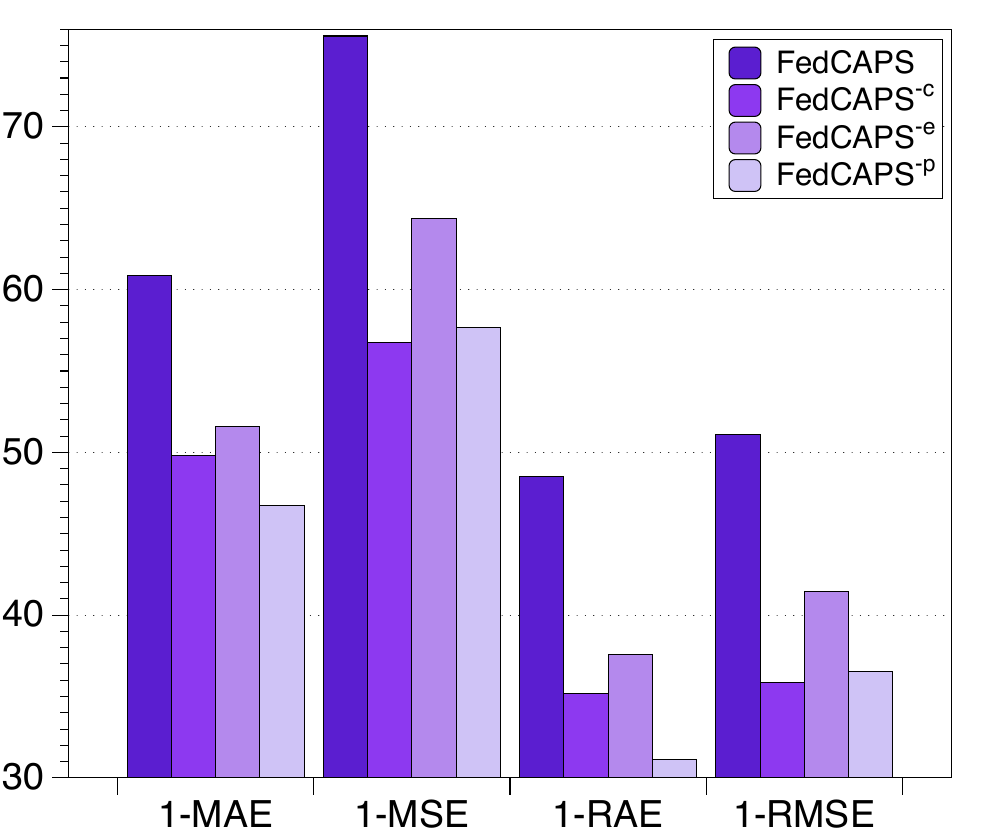}
}
\vspace{-0.1cm}
\caption{The impact of data collection ($^{-c}$), permutation invariance ($^{-e}$) and RL search ($^{-p}$)(\textbf{\model}: (a)(b), \textbf{\extendedmodel}: (c)(d)).}
\vspace{-0.3cm}
\label{abalation_study}
\end{figure*}

\begin{figure*}[!h]
\centering
\subfigure[SpectF (\model)]{
\includegraphics[width=0.23\linewidth, trim = {0 0 1.5cm 0}]{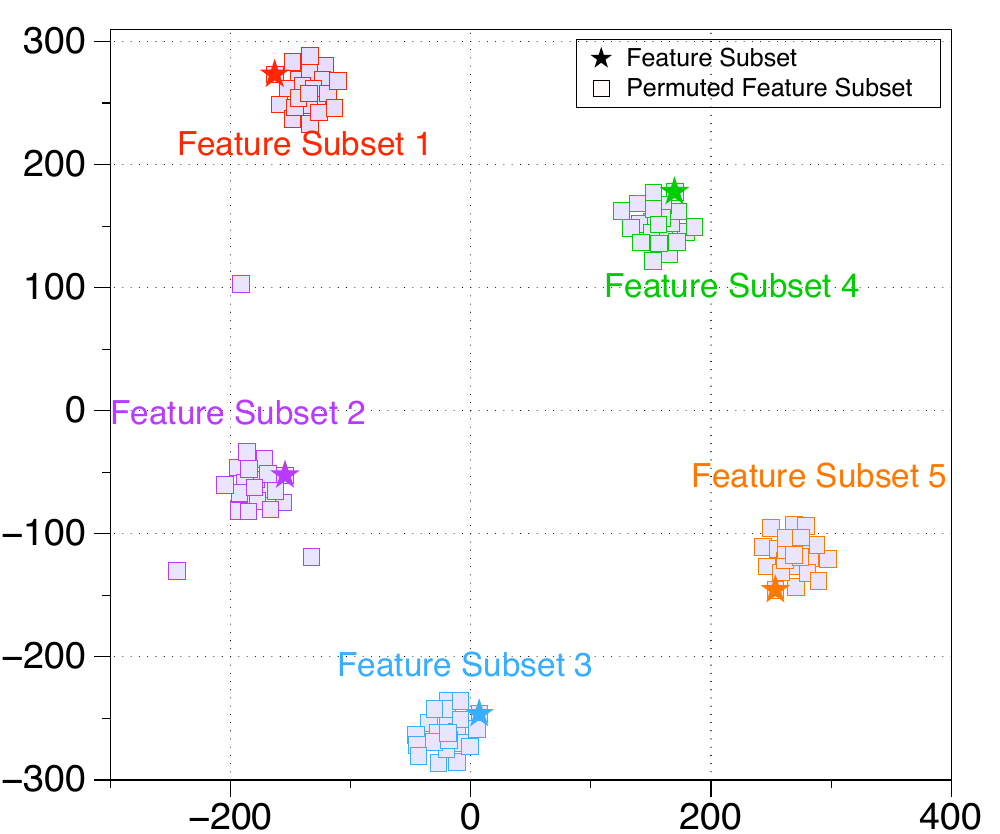}
}
\hspace{-1mm}
\subfigure[German Credit (\model)]{
\includegraphics[width=0.23\linewidth, trim = {0 0 1.5cm 0}]{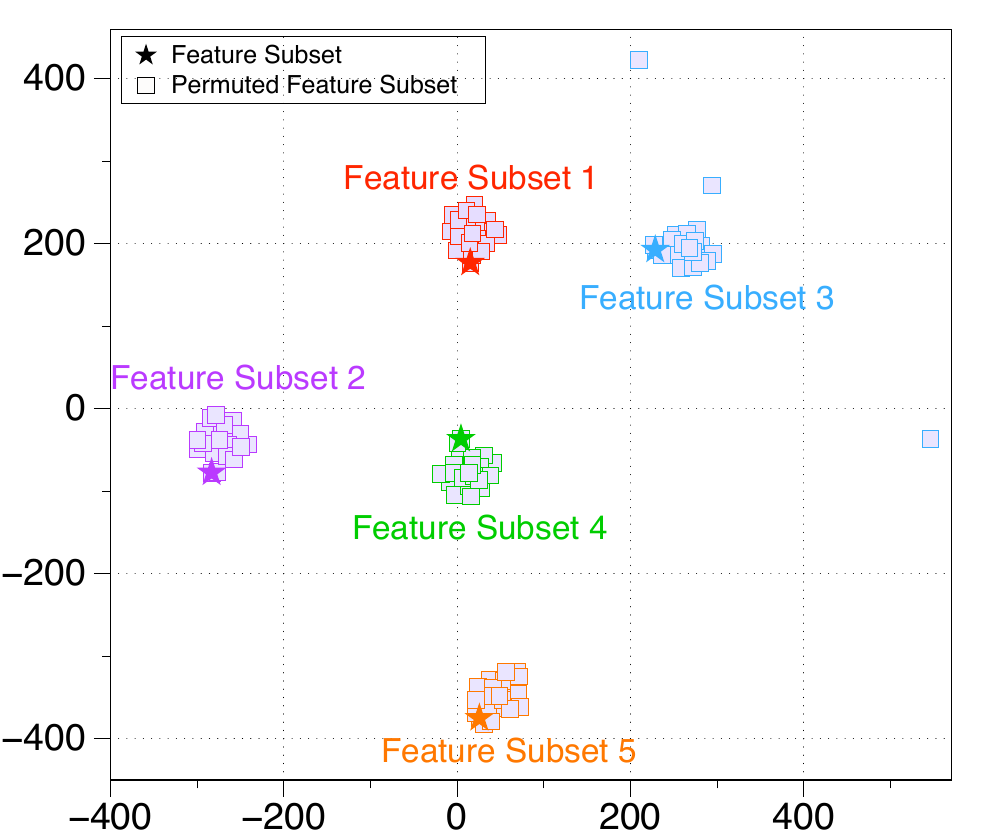}
}
\hspace{-1mm}
\subfigure[SVMGuide3 (\extendedmodel)]{ 
\includegraphics[width=0.23\linewidth, trim = {0 0 1.5cm 0}]{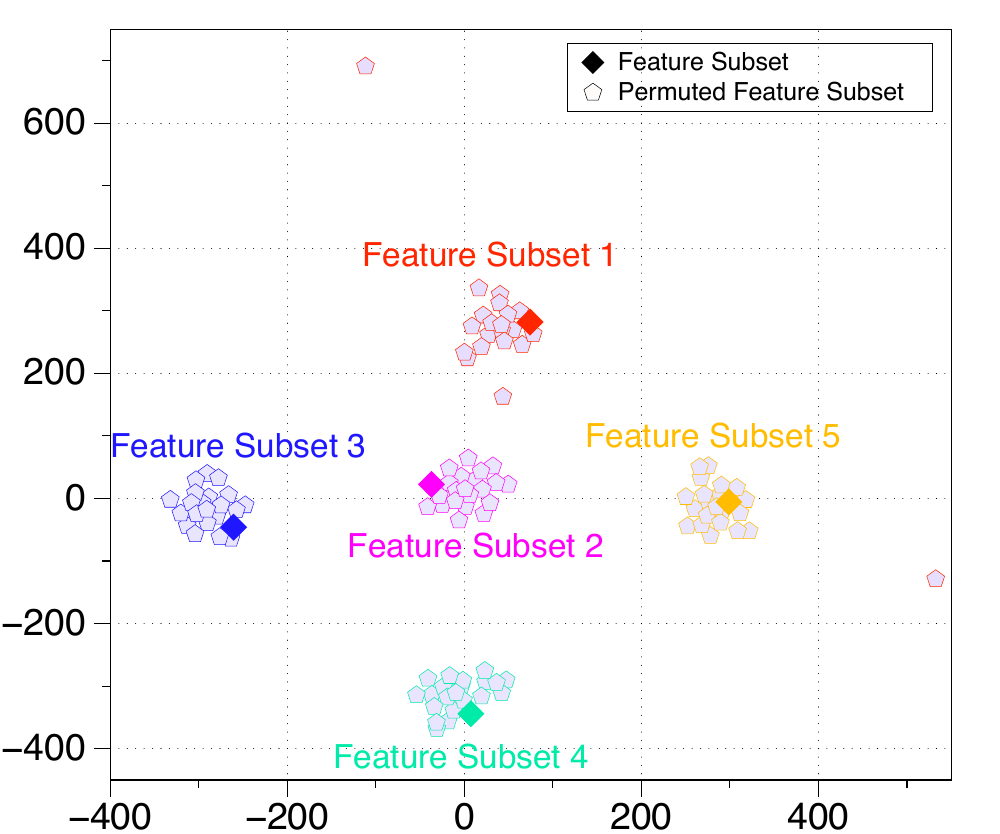}
}
\hspace{-1mm}
\subfigure[Openml\_616 (\extendedmodel)]{ 
\includegraphics[width=0.23\linewidth, trim = {0 0 1.5cm 0}]{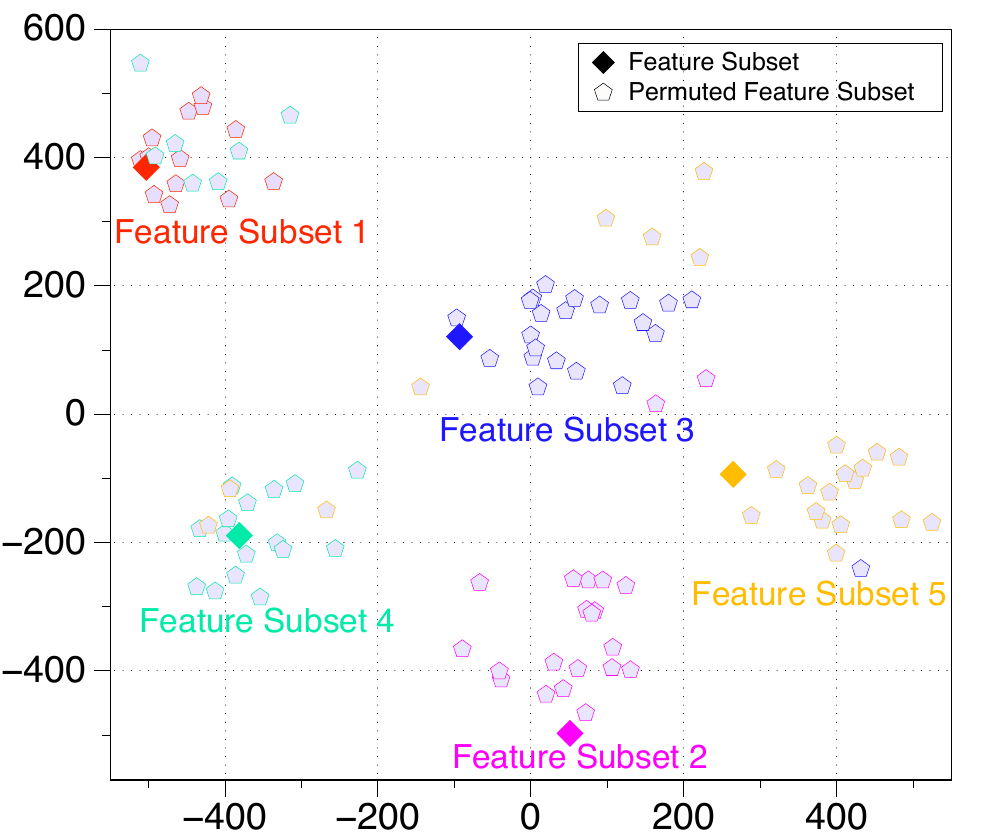}
}
\vspace{-0.1cm}
\caption{The visualization of original and permuted feature subset embeddings (\textbf{\model}: (a)(b), \textbf{\extendedmodel}: (c)(d)).}
\vspace{-0.3cm}
\label{permuted}
\end{figure*}

\subsection{Experimental Setup}
\subsubsection{Dataset Descriptions} 
We evaluate both centralized model \textbf{{\model}} and federated model \textbf{{\extendedmodel}} on 14 publicly accessible datasets from UCI~\cite{uci}, OpenML~\cite{openml}, CAVE~\cite{cave}, Kaggle~\cite{kaggle}, LibSVM~\cite{libsvm}, UrbanSound~\cite{Salamon:UrbanSound:ACMMM:14} and OSF~\cite{osf}. These datasets can be categorized into 3 groups according to the types of ML tasks: 1) binary classification (C); 2) multi-class classification (MC); 3) regression (R). The statistics of these datasets can be found in Table~\ref{table_overall_perf}.

\subsubsection{Evaluation Metrics}
To conduct fair comparisons and alleviate the variance of the downstream ML model, we adopt Random Forest for both \textbf{\model} and \textbf{\extendedmodel} to assess the quality of selected feature subset and report the performance of each baseline algorithm by running five-fold cross-validation method. Moreover, we conduct all experiments via the hold-out setting to further obtain a fair comparison. For binary classification tasks, we use F1-score, Precision, Recall, and ROC/AUC. For multi-classification tasks, we use Micro-F1, Precision, Recall, and Macro-F1. For regression tasks, we use 1-Mean Average Error (1-MAE), 1-Mean Square Error (1-MSE), 1-Relative Absolute Error (1-RAE), and 1-Root Mean Square Error (1-RMSE). 
For all the reported metrics, a higher value consistently indicates better overall model performance.
\subsubsection{Baseline Algorithms}
For centralized model \textbf{\model}, we compare its performance with 12 widely-used feature selection methods: 
(1) \textbf{K-Best}~\cite{kbest} selects top-K features with the best feature scores; 
(2) \textbf{mRMR}~\cite{mrmr} aims to select a feature subset that comprises of the features with greatest relevance and the least redundancy to the target; 
(3) \textbf{LASSO}~\cite{lasso} conducts feature selection by regularizing model parameters, that is shrinking the coefficients of useless features into zero; 
(4) \textbf{RFE}~\cite{rfe} selects features in a recursive fashion, that is removing the weakest features until the stipulated feature numbers is reached; 
(5) \textbf{LASSONET}~\cite{lassonet} conducts feature selection by training neural networks with novel objective function;
(6) \textbf{GFS}~\cite{gfs} conducts feature selection via genetic algorithms that is recursively generating a population based on a possible feature subset first, and then using a predictive model to assess it;
(7) \textbf{SARLFS}~\cite{sarlfs} utilizes an agent to select features among all features to alleviate computational costs and views feature redundancy and downstream task performance as incentives;
(8) \textbf{MARLFS}~\cite{marlfs} is an enhanced version of SARLFS, that is building a multi-agent system to select features with the same incentive settings with SARLFS, and each agent is associated with a single feature;
(9) \textbf{RRA}~\cite{rra} is a rank integration algorithm that first gathers several selected feature subsets, and then integrates them based on the their rank in terms of statistical information;
(10) \textbf{MCDM}~\cite{mcdm} first combines the ranks of features from different feature selection algorithms to form a decision matrix, and then assigns scores to features based on this matrix for the purpose of more comprehensively and accurately selecting the most valuable features through considering multiple evaluation criteria;
(11) \textbf{GAINS}~\cite{gains} first constructs a embedding vector space by jointly optimizing an encoder-evaluator-decoder model, and then uses the gradient from well-evaluator moves the embedding point toward high performance direction to search the optimal feature subset embedding point.
(12) \textbf{MEL}~\cite{mel} divides the parent population into two sub-populations that search for optimal feature subsets independently and uses the mutual influence and knowledge sharing between the sub-populations, integrating multi-task learning with evolutionary learning. Among the introduced baseline algorithms, K-Best and mRMR are categorized as filter methods. Lasso, RFE, and LassoNet are representative of embedded methods. GFS, SARLFS, and MARLFS are wrapper methods. RRA, MCDM, and GAINS are classified as hybrid feature selection methods.

For federated model \textbf{{\extendedmodel}}, we compare its perfromance with 4 widely-used federated learning algorithms:
(1) \textbf{FedAvg}~\cite{fedavg} is a standard baseline in federated learning, where each client performs local updates on its private dataset and sends the resulting model parameters to a central server. The server then updates the global model by computing a weighted average of the client models, where the weight of each client is determined by the ratio of its local sample size to the total sample size of all participating clients.
(2) \textbf{FedNTD}~\cite{fedntd} incorporates a not-true distillation mechanism based on FedAVG, where clients optimize their local models with respect to both the ground-truth labels and additional non-true prediction distribution from the global model, mitigating performance degradation under non-IID data.
(3) \textbf{FedProx}~\cite{fedprox} extends FedAvg by introducing a proximal term into the local objective that penalizes large deviations of client models from the global model. This proximal term improves stability and convergence under heterogeneous data distributions and system conditions while reducing the negative impact of stragglers and unbalanced local updates. 
(4) \textbf{MOON}~\cite{moon} introduces a contrastive loss into the local training objective. The objective aims to alleviate weight divergence by reducing the distance between the representations of local models and global models. It also increases the distance between the representations of a local model and its previous local model to speed up convergence. 

\begin{figure*}[htbp]
    \centering
    \subfigure[RandomForest (\model)]{
        \includegraphics[width = 0.18\linewidth, trim = {0 0 2.5cm 0}]{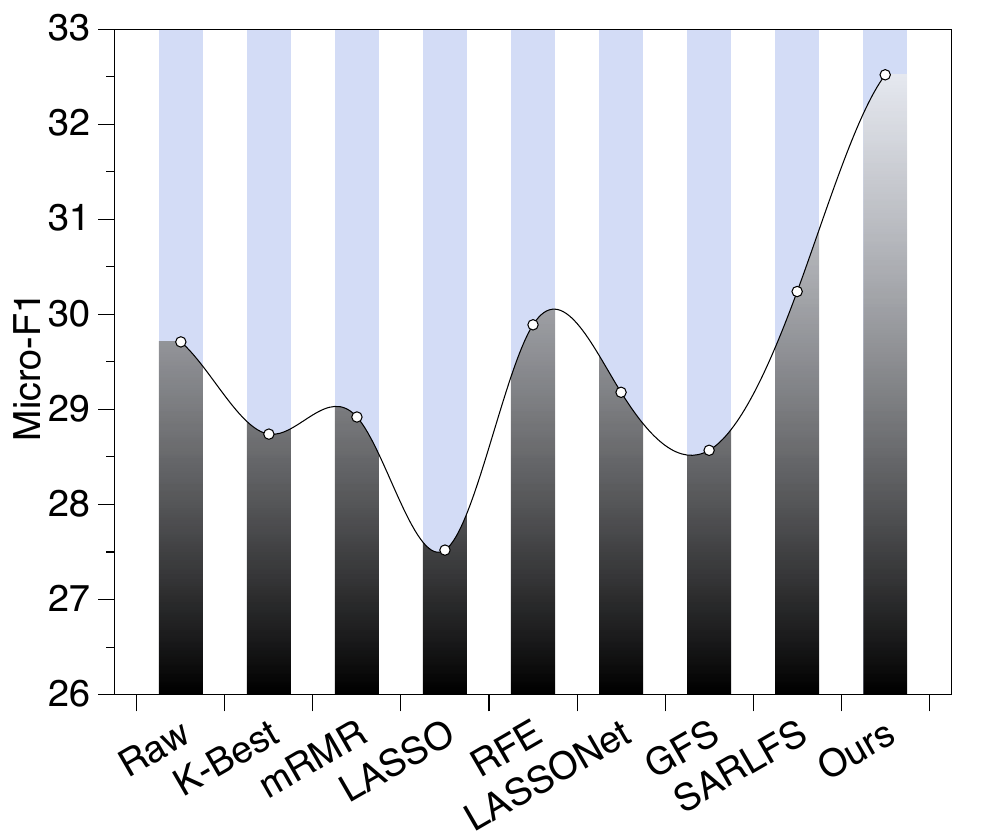}
    }
    \hfill
    \subfigure[XGBoost (\model)]{
        \includegraphics[width = 0.18\linewidth, trim = {0 0 2.5cm 0}]{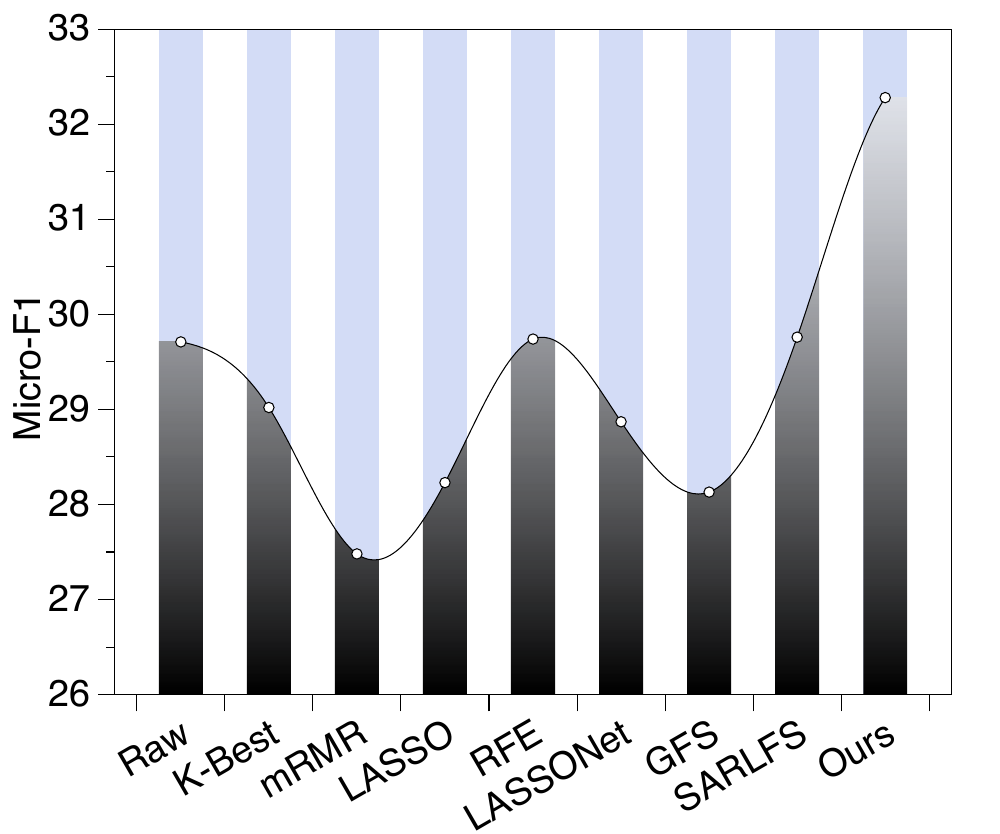}
    }
    \hfill
    \subfigure[SVM (\model)]{
        \includegraphics[width = 0.18\linewidth, trim = {0 0 2.5cm 0}]{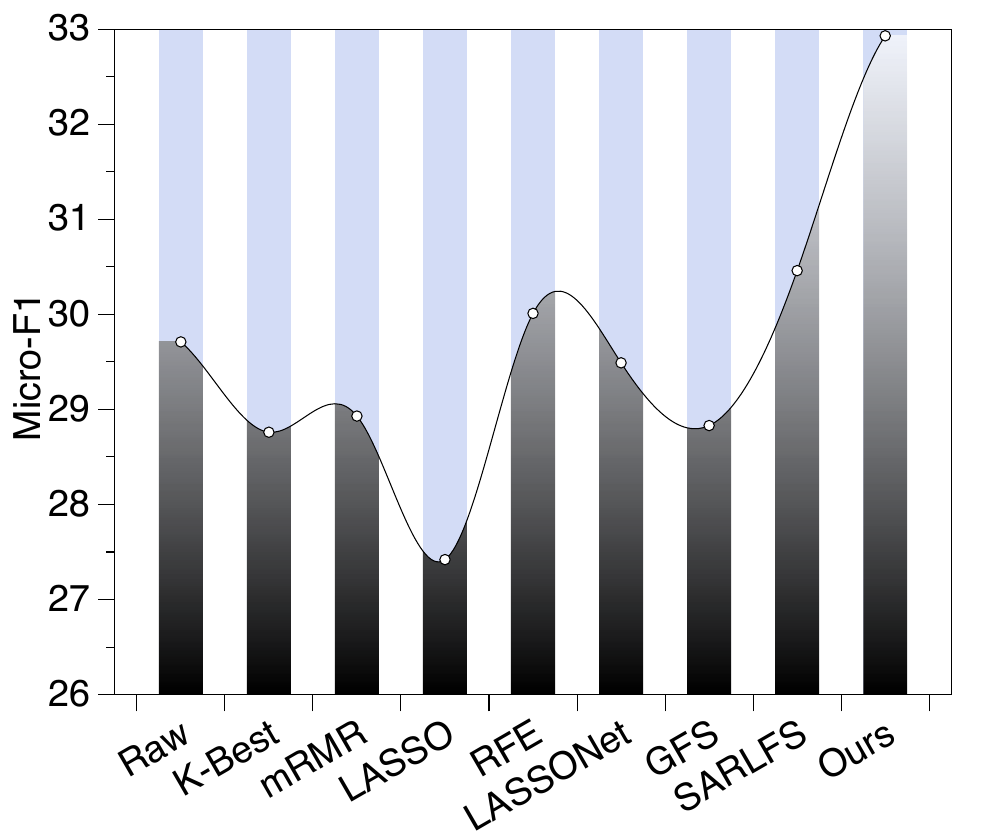}
    }
    \hfill
    \subfigure[KNN (\model)]{
        \includegraphics[width = 0.18\linewidth, trim = {0 0 2.5cm 0}]{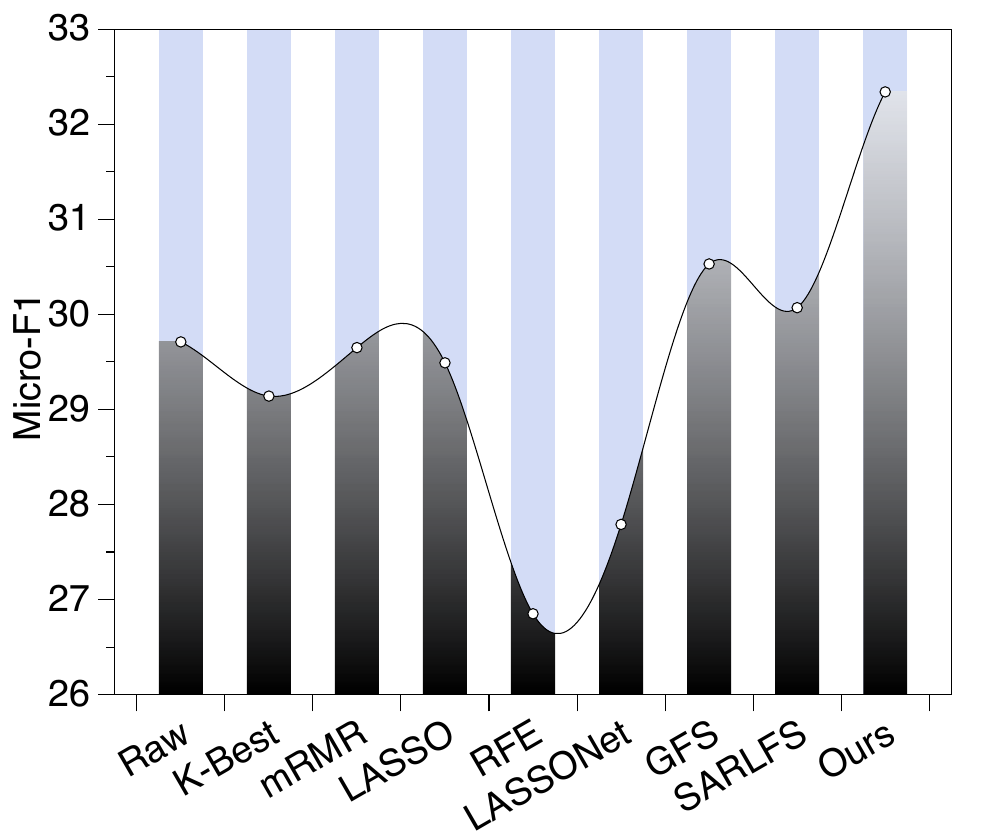}
    }
    \hfill
    \subfigure[Decision Tree (\model)]{
        \includegraphics[width = 0.18\linewidth, trim = {0 0 2.5cm 0}]{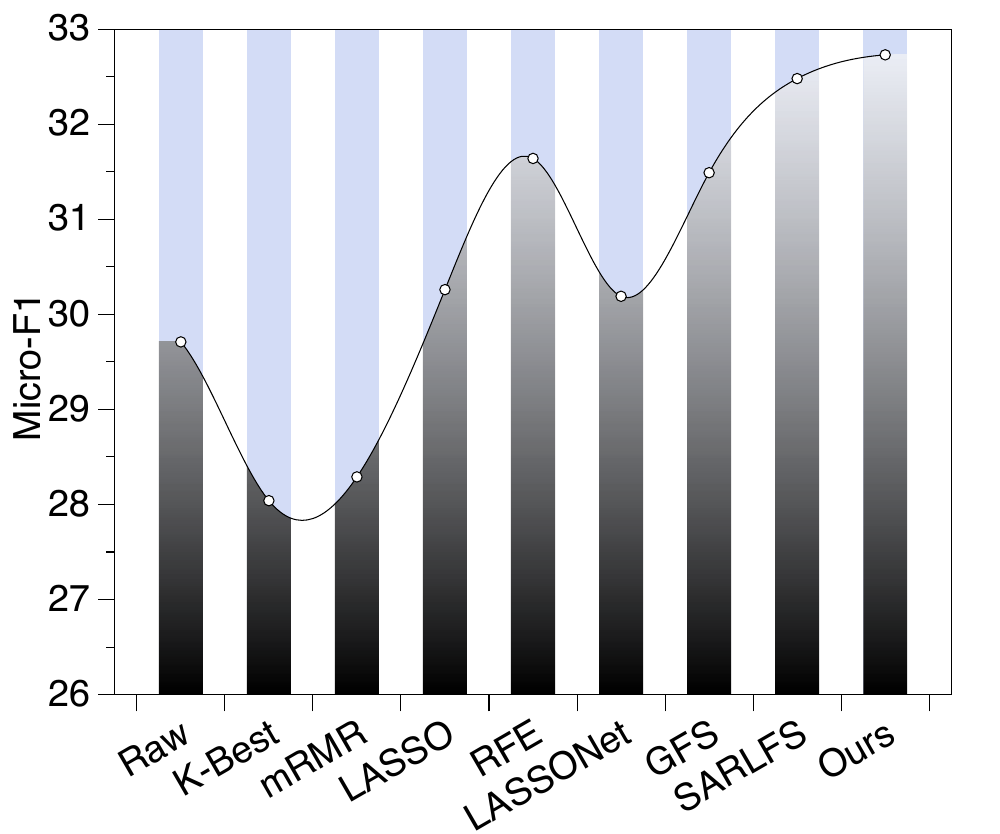}
    }
    \vspace{-0.1cm}
    \caption{Comparison of different downstream ML models in terms of Micro-F1 on UrbanSound (\model).}
    \vspace{-0.3cm}
    \label{table_robust}
\end{figure*}

\begin{figure*}[htbp]
    \centering
    \subfigure[RandomForest (\extendedmodel)]{
        \includegraphics[width = 0.18\linewidth, trim = {0 0 2.5cm 0}]{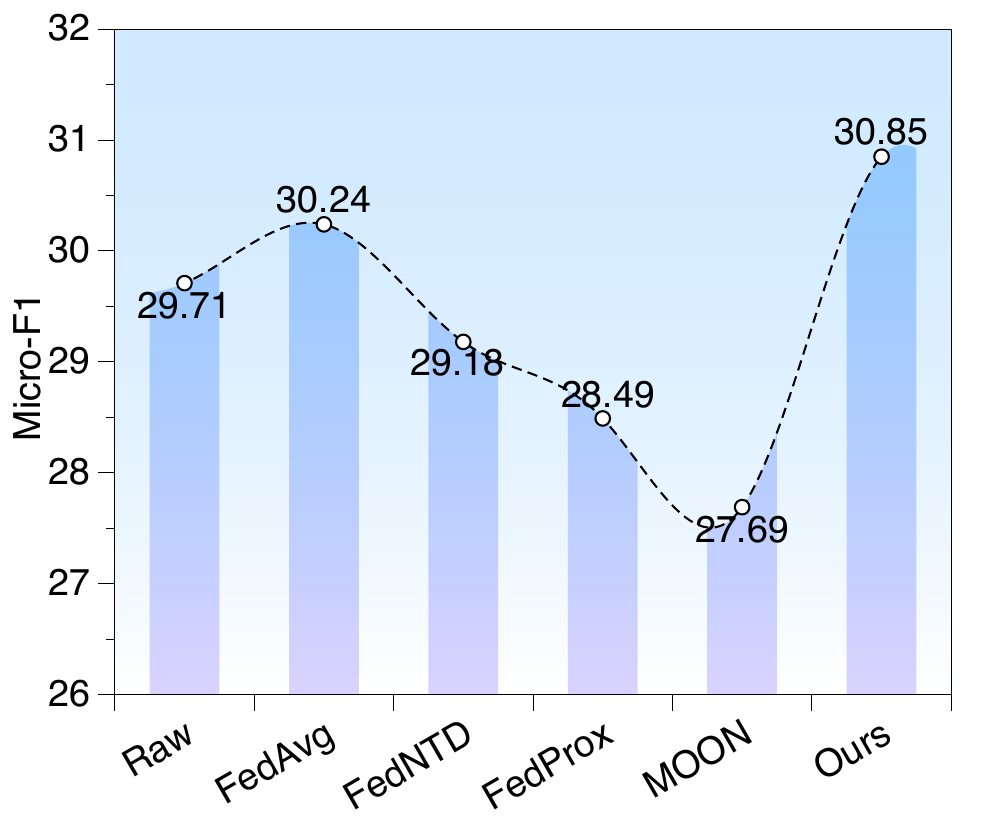}
    }
    \hfill
    \subfigure[XGBoost (\extendedmodel)]{
        \includegraphics[width = 0.18\linewidth, trim = {0 0 2.5cm 0}]{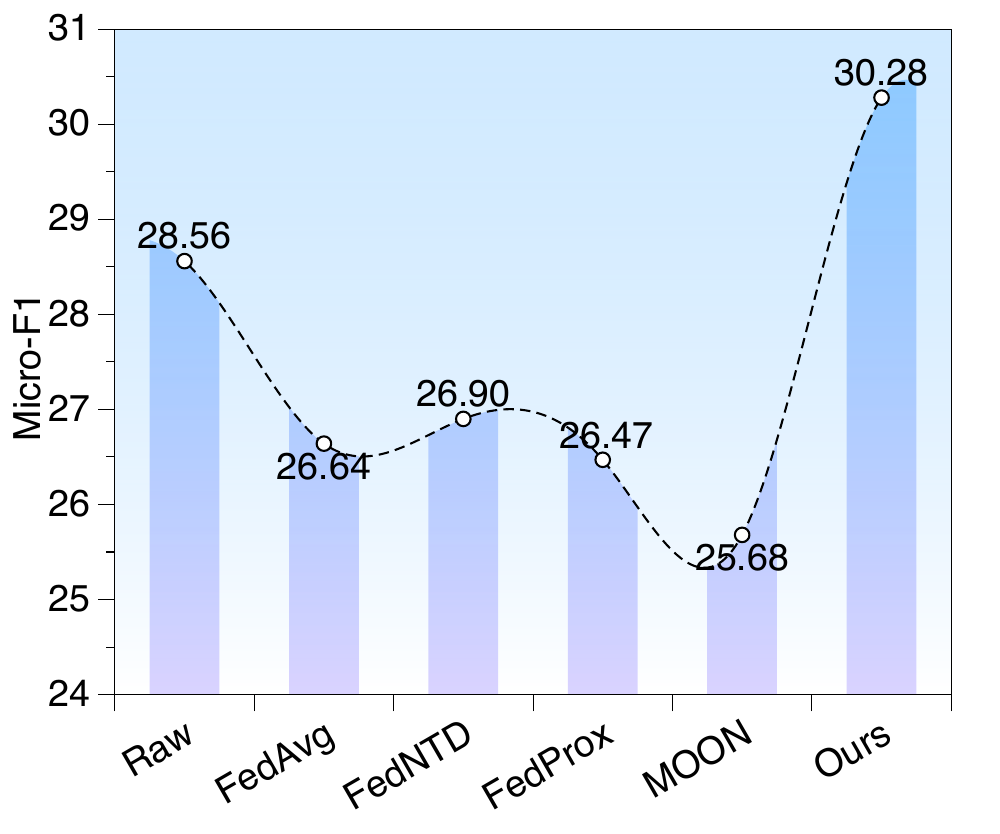}
    }
    \hfill
    \subfigure[SVM (\extendedmodel)]{
        \includegraphics[width = 0.18\linewidth, trim = {0 0 2.5cm 0}]{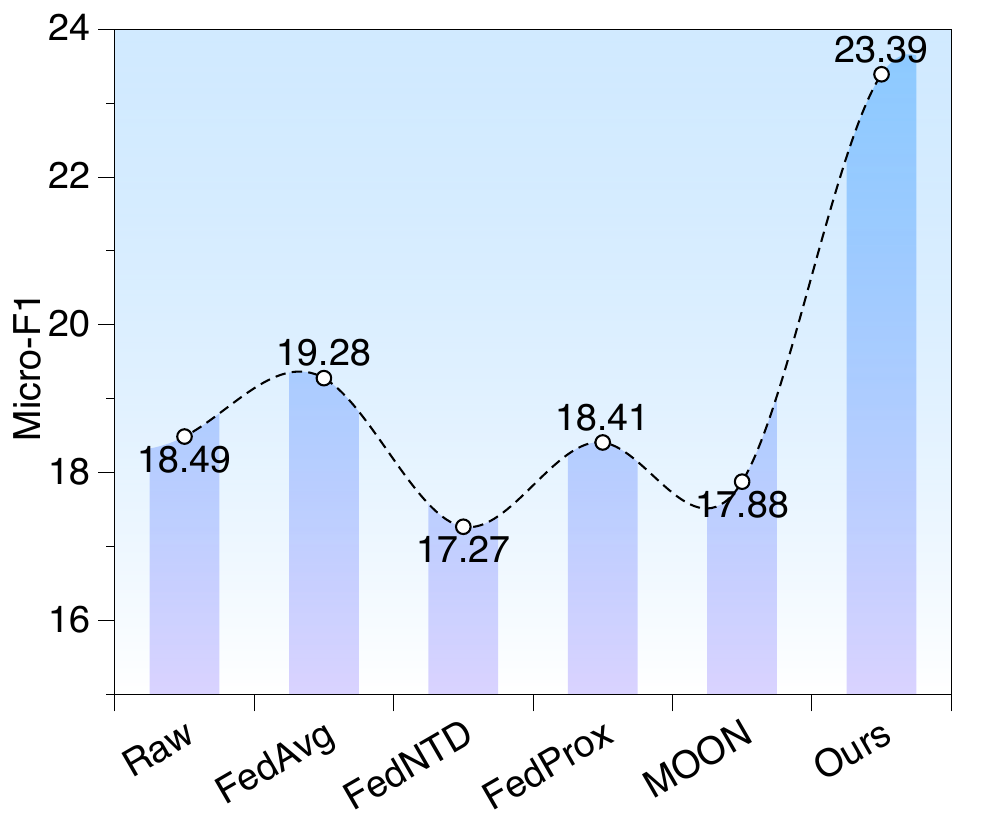}
    }
    \hfill
    \subfigure[KNN (\extendedmodel)]{
        \includegraphics[width = 0.18\linewidth, trim = {0 0 2.5cm 0}]{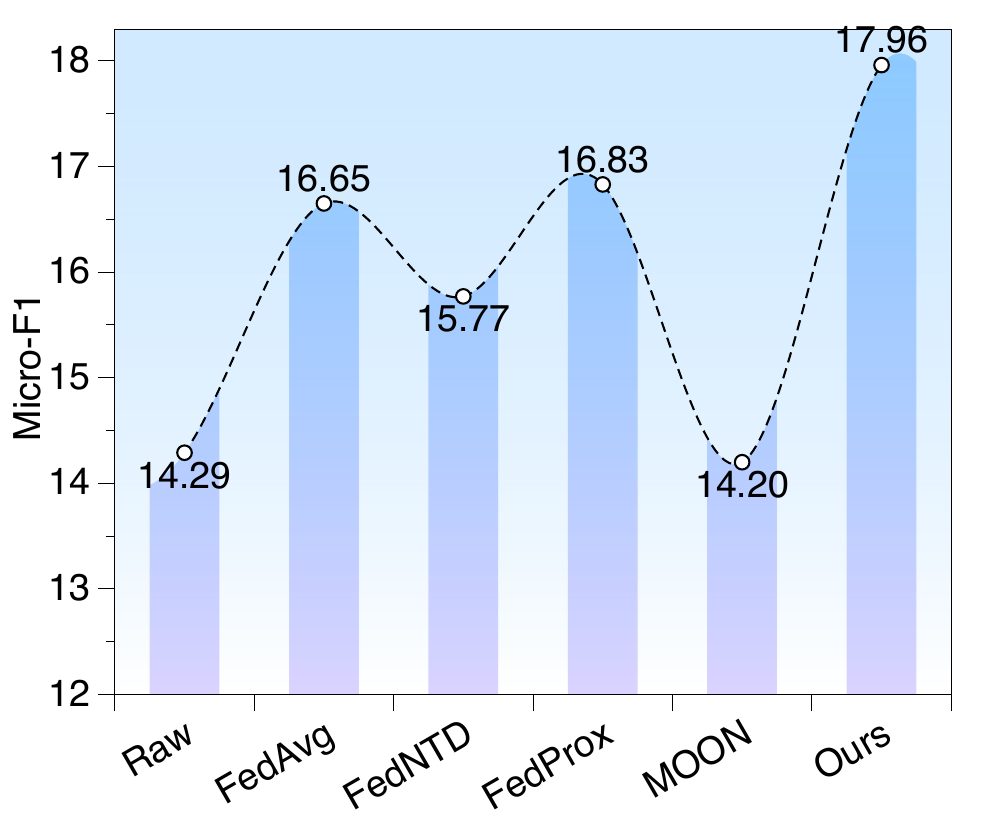}
    }
    \hfill
    \subfigure[Decision Tree (\extendedmodel)]{
        \includegraphics[width = 0.18\linewidth, trim = {0 0 2.5cm 0}]{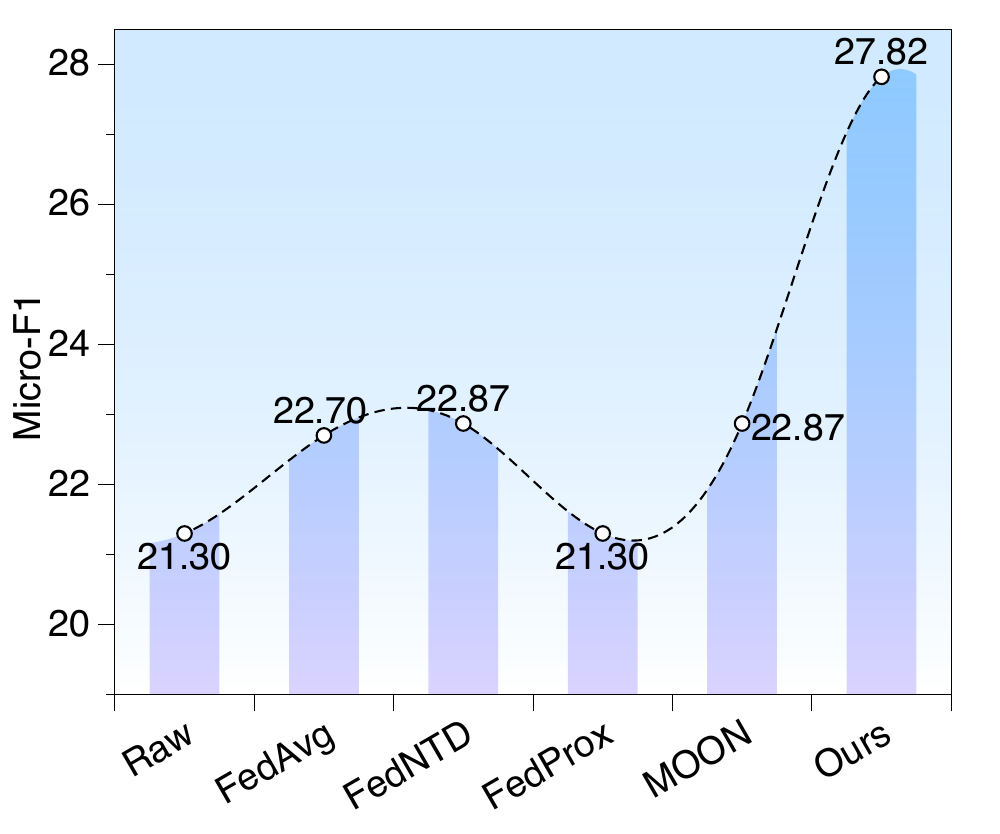}
    }
    \vspace{-0.1cm}
    \caption{Comparison of different downstream ML models in terms of Micro-F1 on UrbanSound (\extendedmodel).}
    \vspace{-0.3cm}
    \label{table_robust_fed}
\end{figure*}

\subsubsection{Experimental Settings}
We conducted all experiments on the Windows 11 operating system, AMD Ryzen 5 5600X CPU, and NVIDIA GeForce RTX 3070Ti GPU. 
We employ the framework of Python 3.10.15 and PyTorch 2.5.1~\cite{pytorch}.

\subsubsection{Hyperparameters and Reproducibility for \textbf{{\model}} and \textbf{{\extendedmodel}}}
To collect potential feature subsets and their corresponding model performance, we run MARLFS for 300 epochs. 
To increase the diversity of training data, we augment the collected feature subset-accuracy records by permuting the order of each feature subset indices 25 times. 
The Encoder comprises of two layers of ISAB. 
The Decoder comprises of 3 components, which are PMA, MAB and rFF. 
The number of multi-attention head is 4. 
The embedding size of each feature index is 128. 
To train the encoder and decoder, we set batch size as 64, the step size as 0.001, and the dimension of seed vectors and inducing points as 32 respectively. 
During the search process, we used the top 25 feature selection records as the initial search seeds to search for the optimal embedding vector. 
We set search epoch as 10, batch size as 512, learning rate of the actor as 0.0003, learning rate of the critic as 0.001, reward trade-off as 0.1, reward discounted factor as 0.99, search step as 1000, and the clipping ratio as 0.2, respectively.

\begin{figure*}[!h]
\vspace{-0.2cm}
\centering
\subfigure[F1 Score (\model)]{ 
\includegraphics[width=0.23\linewidth, trim = {0 0 1.5cm 0}]{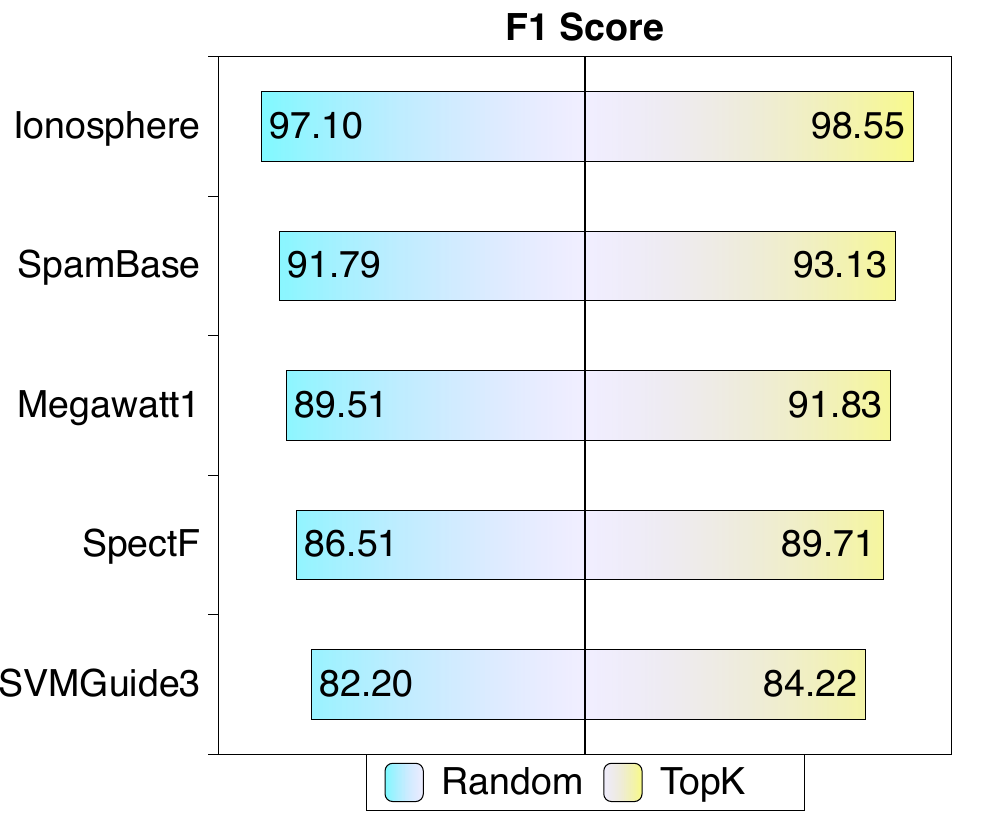}
}
\hspace{-1mm}
\subfigure[ROC/AUC (\model)]{ 
\includegraphics[width=0.23\linewidth, trim = {0 0 1.5cm 0}]{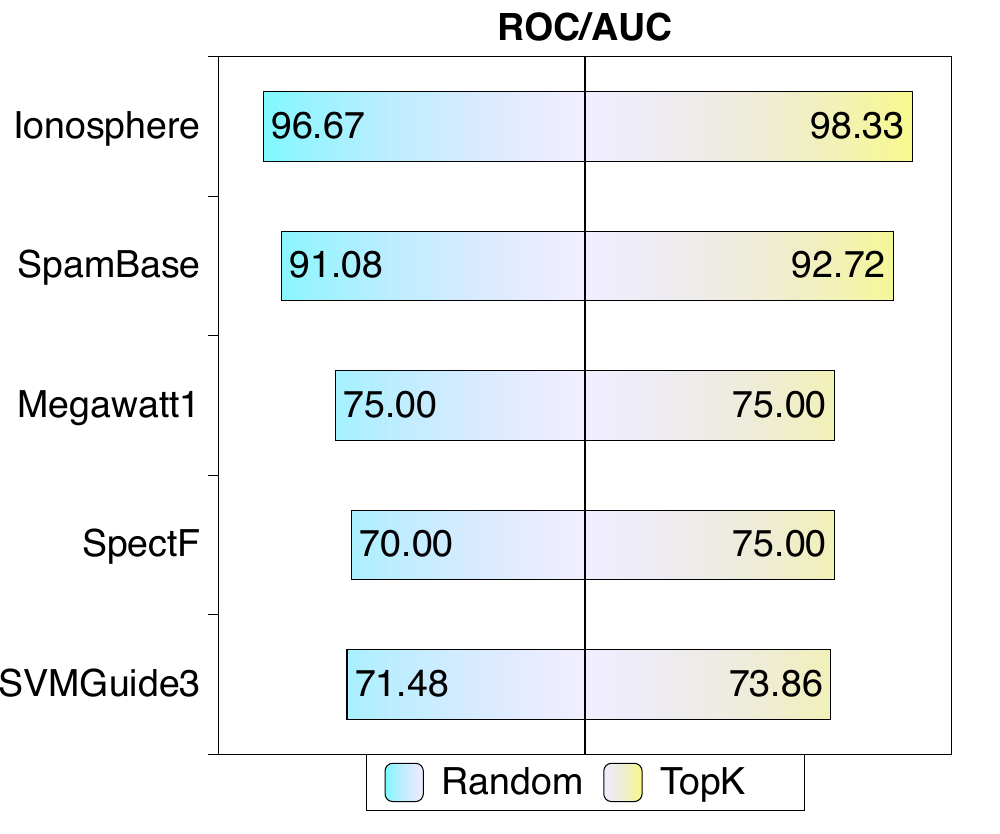}
}
\hspace{-1mm}
\subfigure[F1 Score (\extendedmodel)]{ 
\includegraphics[width=0.23\linewidth, trim = {0 0 1.5cm 0}]{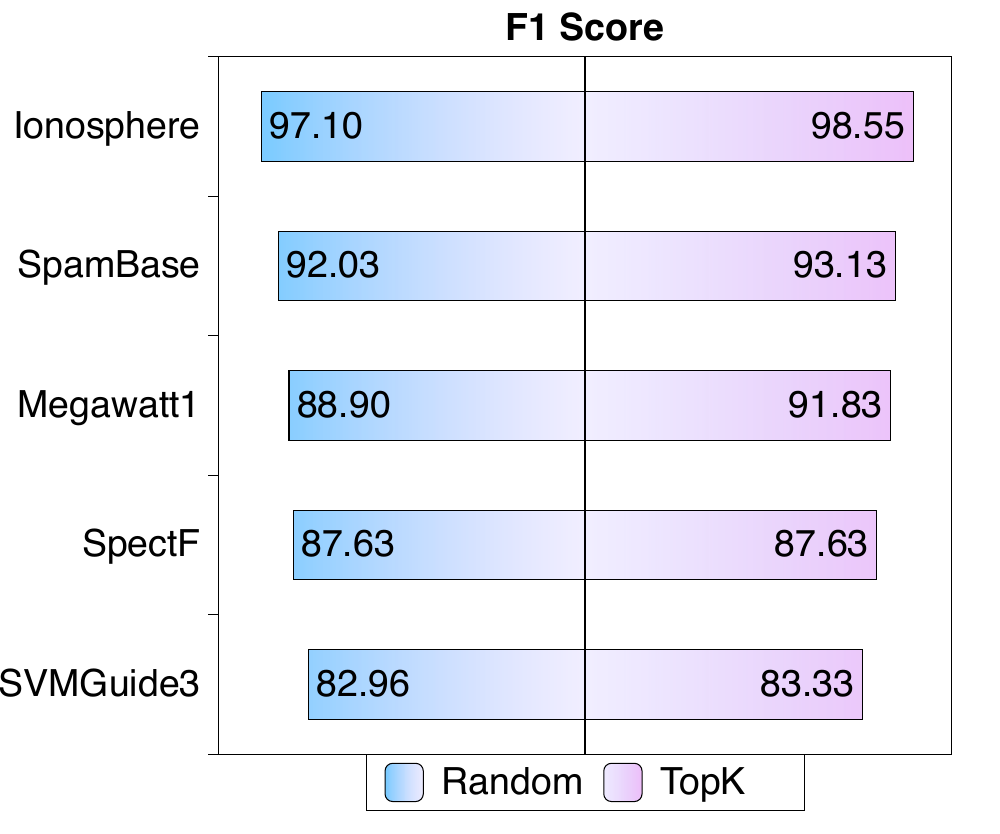}
}
\hspace{-1mm}
\subfigure[ROC/AUC (\extendedmodel)]{ 
\includegraphics[width=0.23\linewidth, trim = {0 0 1.5cm 0}]{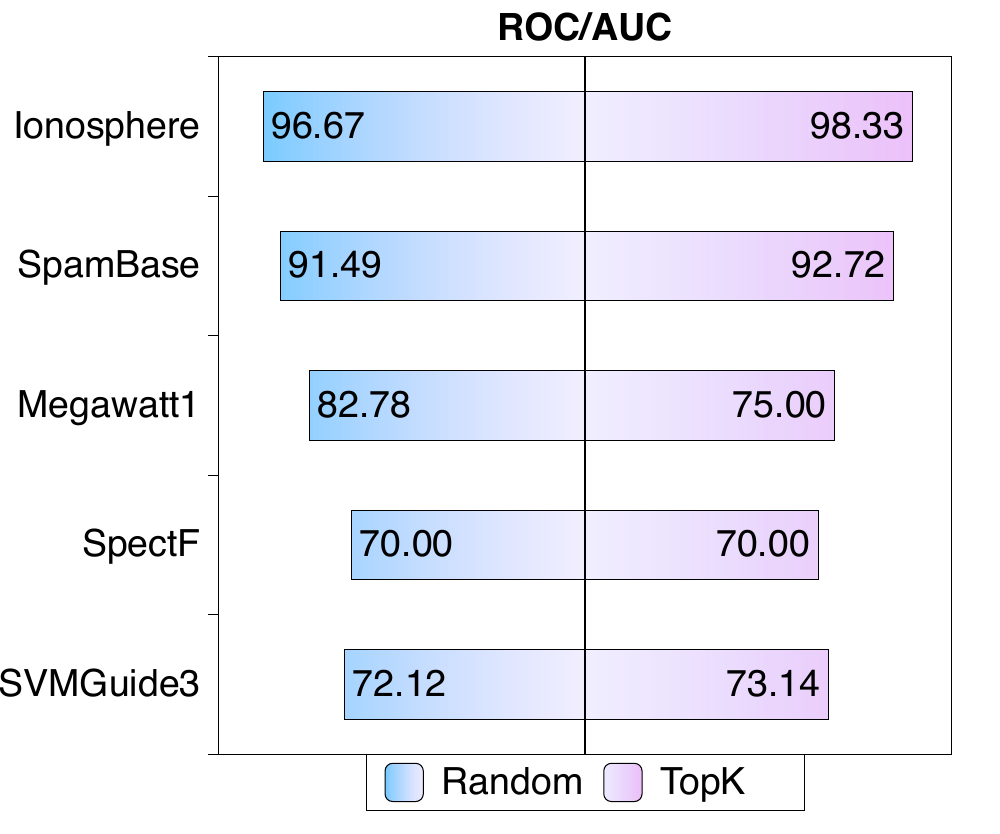}
}
\vspace{-0.1cm}
\caption{Comparison of random search seeds and top-K search seeds in terms of F1 Score and ROC/AUC.}
\vspace{-0.3cm}
\label{search_seeds}
\end{figure*}

\vspace{-0.25cm}
\subsection{Experimental Results}
\subsubsection{Overall Performance of \textbf{{\model}} and \textbf{{\extendedmodel}}}
This experiment is conducted to evaluate the performance of both the centralized model {\model} and federated model {\extendedmodel}. 
\begin{itemize}
    \item Table~\ref{table_overall_perf} shows the comparison of our centralized model \textbf{\model} and 12 baseline feature selection algorithms in terms of F1-score, Micro-F1, and 1-RAE. 
    \item Table~\ref{table_fed_methods} illustrates the comparison of our federated model \textbf{\extendedmodel} and 4 baseline federated learning algorithms in terms of F1-score, Micro-F1, and 1-RAE. 
\end{itemize}
The results demonstrate that our method consistently outperforms all baseline algorithms in the centralized setting, while in the federated setting it achieves the best performance on most datasets and ranks second on the Credit Default dataset. 
There are four possible reasons for this observation: 
a) The permutation-invariant encoder-decoder framework preserves feature subset knowledge in the learned embedding space, focusing on feature pairwise interactions rather than their order and removing the permutation bias in the embedding space;
b) The policy-based RL agent effectively explores the learned continuous embedding space and identifies the superior feature subset embeddings, overcoming the challenges of the non-convex space and converging to the global optimal feature subset embedding.
c) The sample-aware weighted aggregation strategy ensures that clients with larger datasets contribute proportionally more to the global update, reducing the noise from small-sample clients. 
d) The knowledge aggregation preserves more reliable selection–performance relations and therefore establishes a more robust unified embedding space.

\vspace{0.05cm}
\subsubsection{Ablation Study of \textbf{\model} and \textbf{\extendedmodel}}
We conduct this experiment to study the impact of the data collection, permutation-invariant embedding, and policy-guided search on the model performance in both centralized and federated scenarios. 
We implement six model variants: 
\begin{itemize}
    \item \textbf{\model$^{-c}$} and \textbf{\extendedmodel$^{-c}$} collect the feature training records randomly instead of using RL-based collector.
    \item \textbf{\model$^{-e}$} and \textbf{\extendedmodel$^{-e}$} utilize sequential encoder-decoder model to learn the continuous embedding space.
    \item \textbf{\model$^{-p}$} and \textbf{\extendedmodel$^{-p}$} replace policy-guided search with Genetic Algorithm (GA) to explore the embedding space and identify the optimal feature subset embedding.
\end{itemize}
We randomly choose four datasets to conduct this experiments. SVMGuide3 and German Credit are used to evaluate the performance of the centralized version, while Mice Protein and Openml\_616 are employed to evaluate the federated version.
For each type of task, we evaluate the model performance from four different perspectives. For binary classification tasks, we use Precision, Recall, F1-score, and ROC/AUC. 
For multi-classification tasks, we use Precision, Recall, Micro-F1, and Macro-F1. 
For regression tasks, we use 1-MAE, 1-MSE, 1-RAE, and 1-RMSE. 
Figure~\ref{abalation_study} shows the overall experiment results. 
We observe that the original version consistently surpasses its variants ($^{-c}$, $^{-e}$, and $^{-p}$) across all datasets in both centralized and federated scenarios.
The potential reasons for this observation are: 
1) the feature selection records generated by the data collectors are more accurate and robust, constructing a more comprehensive and effective continuous embedding space for searching the optimal feature subset; 
2) sequential encoder-decoder model fails to embed permutation invariance in the embedding space, introducing permutation bias and leading to local optima; 
3) policy-guided RL agent conducts effective exploration for optimal feature subset, eliminating the reliance on convexity assumptions and avoiding convergence to local optima.
Thus, this experiment illustrates the significance of data collector, permutation-invariant embedding and policy-guided search in enhancing the model performance.

\begin{figure}[!t]
    \centering
    \includegraphics[width=\linewidth]{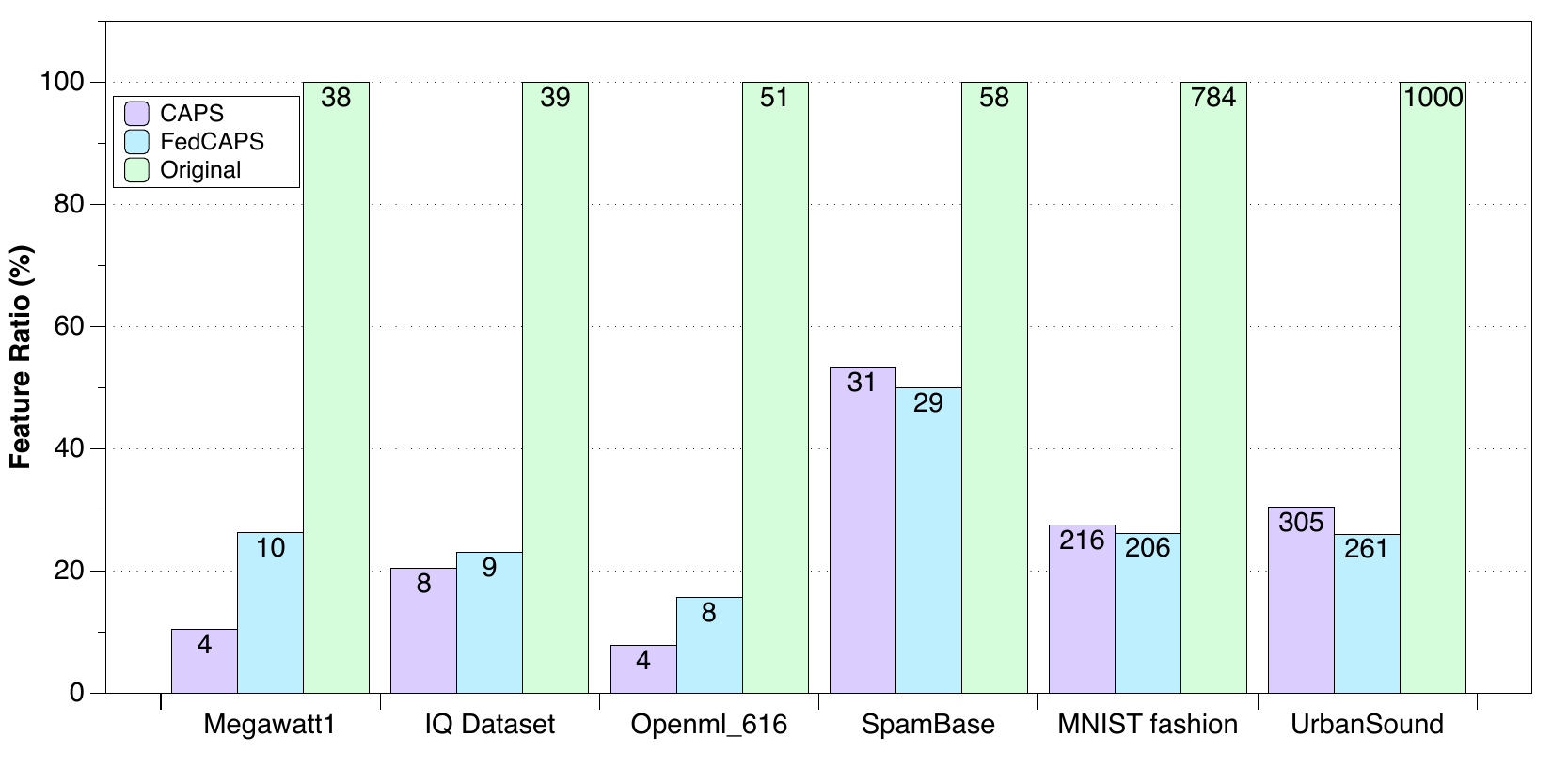}
    \vspace{-0.1cm}
    \caption{Comparison of the selected feature subset size of centralized model (\model), the federated model(\extendedmodel), and the original feature set (Original).}
    \label{feat_num}
    \vspace{-0.3cm}
\end{figure}

\subsubsection{Study on the permutation sensitivity of learned feature subset embeddings (\textbf{\model} and \textbf{\extendedmodel})} 
We conduct this experiment to study the permutation sensitivity of learned feature subset embeddings. We randomly choose four datasets and visualize the embeddings of the original feature subsets along with their corresponding permuted versions. In detail, we first randomly collect five distinct feature subset as the original representatives. Each feature subset is subjected to a permutation operation for 20 times to generate the corresponding permuted representatives. 
Then, we employ the well-trained encoder to obtain the embeddings for both the original and permuted feature subsets. Subsequently, we use T-SNE to map these embeddings into a 2-dimensional space for visualization. 
Figure~\ref{permuted} shows the visualization results, in which each color represents a unique feature subset along with its corresponding permuted versions. 
The figure contains four subplots, where the left two present the visualization results for the SpectF and German Credit datasets in the centralized scenario, and the right two show the visualization result for the SVMGuide3 and Openml\_616 datasets in the federated scenario.
The embeddings of the original feature subset and the corresponding permuted versions are represented by pentagrams and squares respectively. 
We find that the distribution locations of the permuted feature subset embeddings are consistently clustered around their corresponding original feature subset embedding. 
A potential reason for this observation is that the learned continuous embedding space inherently captures permutation invariance, ensuring that variations in feature order do not impact the encoded embedding distribution. Therefore, this experiment demonstrates that our proposed encoder-decoder embeds 
feature subset knowledge into a permutation-invariant embedding space, successfully removing the permutation bias.

\begin{figure}[!t]
        \centering
        \subfigure[CAPS]{
        \includegraphics[width=0.46\linewidth,trim = {3cm 1.5cm 3cm 2.5cm}]{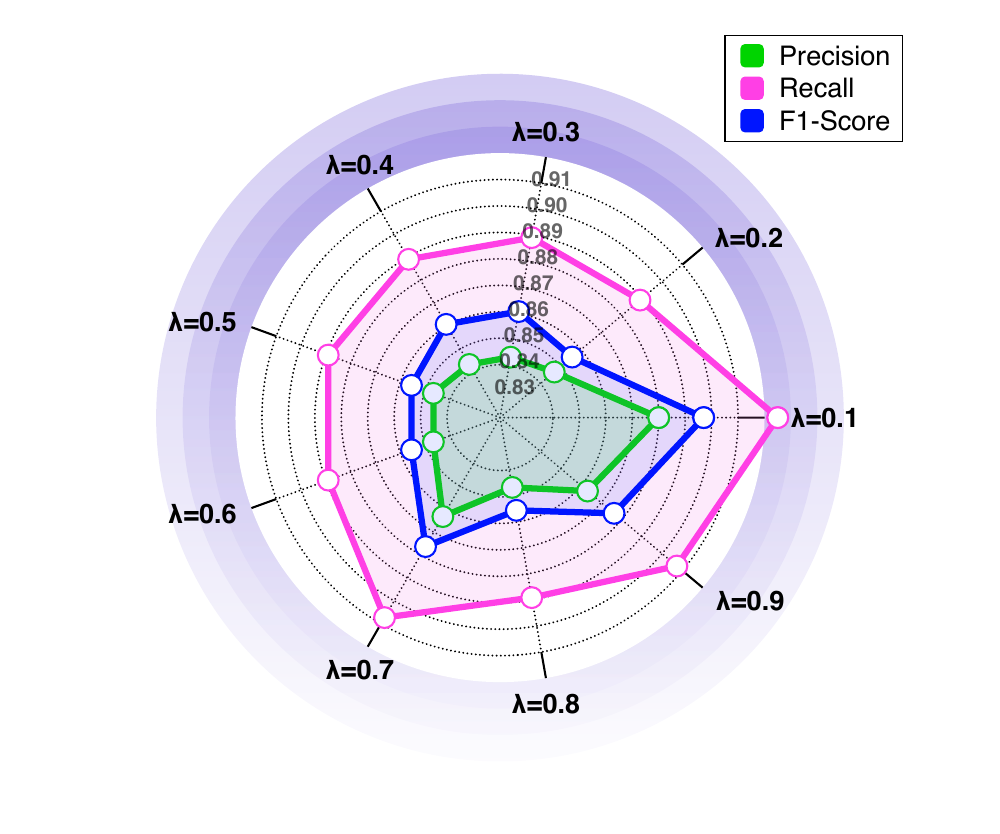}
        }
        \hfill
        \subfigure[FedCAPS]{
        \includegraphics[width=0.46\linewidth,trim = {3cm 1.5cm 3cm 2.5cm}]{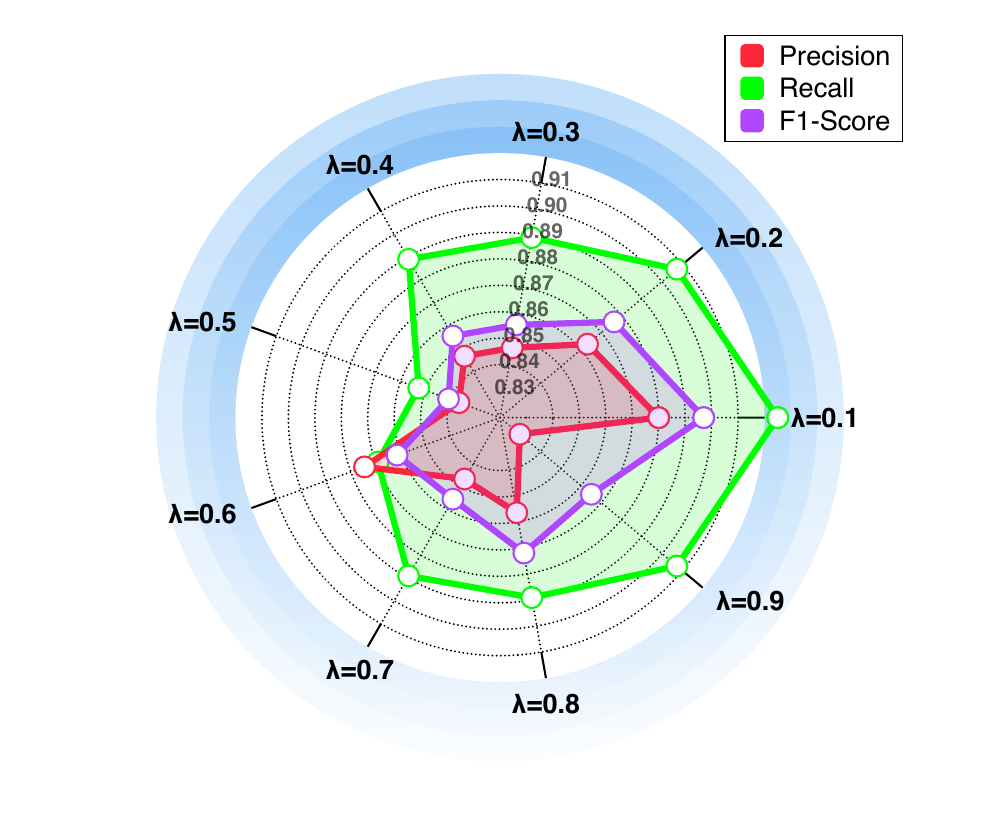}
        }
        \vspace{-0.1cm}
        \caption{The hyperparameter sensitivity test on SpectF dataset.}
        \label{trade}
        \vspace{-0.3cm}
\end{figure}

\subsubsection{Study on the impact of search seeds in \textbf{\model} and \textbf{\extendedmodel}}
We conduct this experiment to study the impact of initial search seeds for the RL-based search process in both centralized and federated scenarios. 
To mitigate bias, we conduct the comparison on five randomly chosen datasets and in term of two metrics, which are F1 Score, and ROC/AUC. 
We compare the search process using random K historical records with the one using top-K historical records based on model performance. 
Figure~\ref{search_seeds} shows the overall comparison results. 
The left two subplots present the results in the centralized scenario, while the right two subplots present the results in the federated scenario.
The experimental results demonstrate that the search process with top-K records as initial search seeds consistently outperforms the one without across all metrics. 
A plausible explanation for this observation is that the starting of the search process is sensitive to the initial starting points. Initializing the search with top-K historical records not only provides a more promising starting region in the embedding space but also reduces the likelihood of exploring less relevant or suboptimal areas. The RL-based search agent can leverage the historical information to form a lower bound and focus on refining already competitive solutions. This leads to faster convergence and more stable performance across various datasets. Another noteworthy observation is that using top-K seeds helps to alleviate the risk of becoming trapped in local optima since these seeds often span multiple good regions in the embedding space. 
In contrast, random seeds might guide the search process toward less advantageous regions, increasing the variance in final performance. 
Thus, these observations show that search seeds are crucial for RL agent to explore the space and identify optimal feature subset.

\subsubsection{Study on the robustness of \textbf{{\model}} and \textbf{{\extendedmodel}} over downstream ML tasks.}
We conduct this experiment to assess the robustness of {\model} and {\extendedmodel} by changing the downstream ML model to Random Forest (RF), XGBoost (XGB), Support Vector Machine (SVM), K-Nearest Neighborhood (KNN), and Decision Tree (DT) respectively. 
Figure~\ref{table_robust} and \ref{table_robust_fed} show the comparison results on UrbanSound dataset in terms of Micro-F1. 
The results demonstrate that {\model} and {\extendedmodel} outperform other baselines across downstream ML models in both centralized and federated scenarios.
A potential reason for this observation is that the data collector can customize the feature selection records based on the downstream ML task. 
Then, the encoder-decoder model is capable of embedding the preference and properties of the ML model into the continuous embedding space. 
Finally, by leveraging these information, the RL agent can reach the global optimal embedding point instead of local optima. 
Thus, this experiment shows the overall robustness of both {\model} and {\extendedmodel}.

\subsubsection{Study on the size of selected feature subsets in \textbf{{\model}} and \textbf{{\extendedmodel}}}
We conduct this experiment to assess the ability of the RL agent to optimize multiple objectives simultaneously, improving model performance while reducing selected features.
We compare the size of the feature subsets selected by our centralized and federated models with the size of the original feature set. 
Figure~\ref{feat_num} demonstrates the results of {\model} and {\extendedmodel} on six randomly chosen datasets in terms of the ratio of selected feature subset size to original feature set size. 
The ratio indicates the proportion of selected features relative to the original features. 
The number associated with each bar is the exact length of selected feature subset.
The results show that the selections of our methods are significantly smaller than the original dataset. 
Although our selected feature subsets contain noticeably fewer features, they consistently achieve superior or comparable model performance. 
A potential reason for this observation is that the policy-guided RL agent is capable of optimizing the learned embeddings by maximizing the downstream task performance and minimizing the length of the selected subset. 
Therefore, this experiment successfully demonstrates that the policy-guided RL agent excels at striking a balance between model performance and feature-efficiency.

\subsubsection{Study on the Hyperparameter Sensitivity of Search Process in \textbf{{\model}} and \textbf{{\extendedmodel}}}
We conduct this experiment to study the sensitivity of the search process performance to the reward trade-off $\lambda$.
The reward trade-off $\lambda$ is designed for the RL agent to balance the attention on the feature subset performance and the total length of the subset from equation. 
In particular, a larger $\lambda$ prioritizes the feature subset performance over total length, and vice versa. 
We vary $\lambda$ from 0.1 to 0.9 and train both the centralized model {\model} and federated model {\extendedmodel} on Spectf dataset. 
The results are shown in Figure~\ref{trade} in terms of Precision, Recall and F1-Score. 
The left figure demonstrates the performance of centralized model \model. 
The right figure illustrates the performance of federated model \extendedmodel.
Overall, we observed that the model performance remains relatively stable across most values of $\lambda$ and occasionally increases when $\lambda$ is set to 0.1. 
These findings provide insights into how $\lambda$ effects the results and how we should choose optimal values for them.

\subsubsection{Case study on \textbf{{\model}} and \textbf{{\extendedmodel}}}
\begin{figure}[!t]
        \centering
        \subfigure[{\model}-selected Feature Subset]{
        \includegraphics[width=0.46\linewidth,trim=0.5cm 0.3cm 0.5cm 0.3cm, clip]{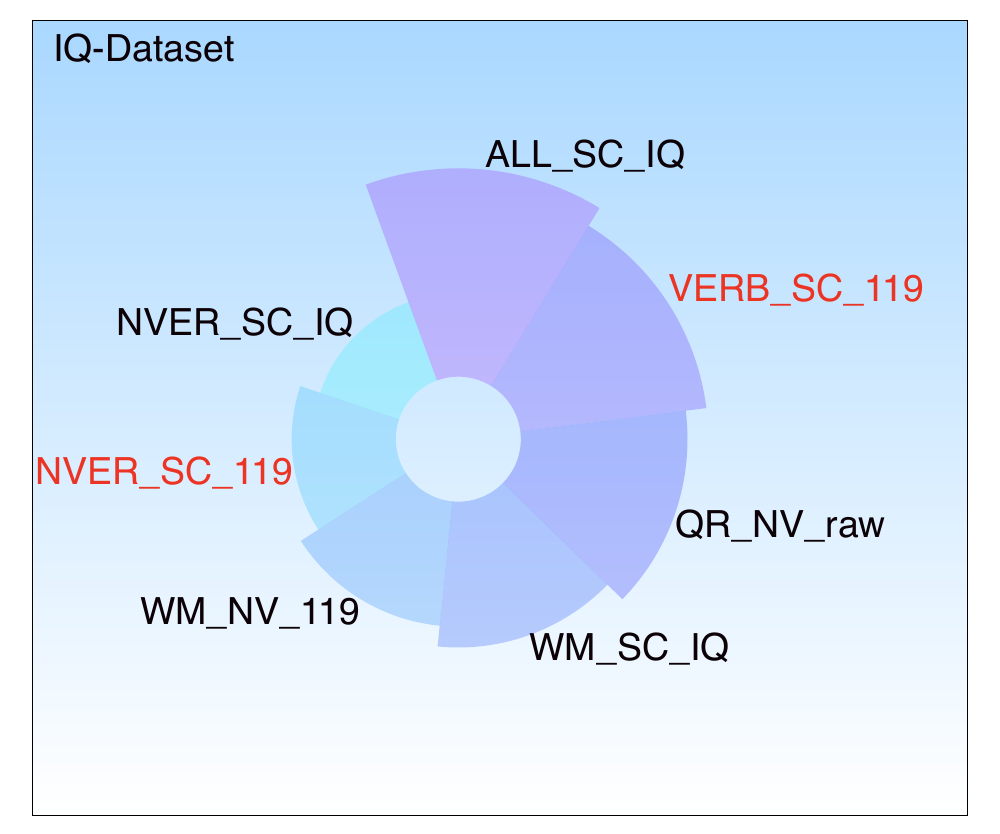}
        }
        \hfill
        \subfigure[{\extendedmodel}-selected Feature Subset]{
        \includegraphics[width=0.46\linewidth,trim=0.5cm 0.3cm 0.5cm 0.3cm, clip]{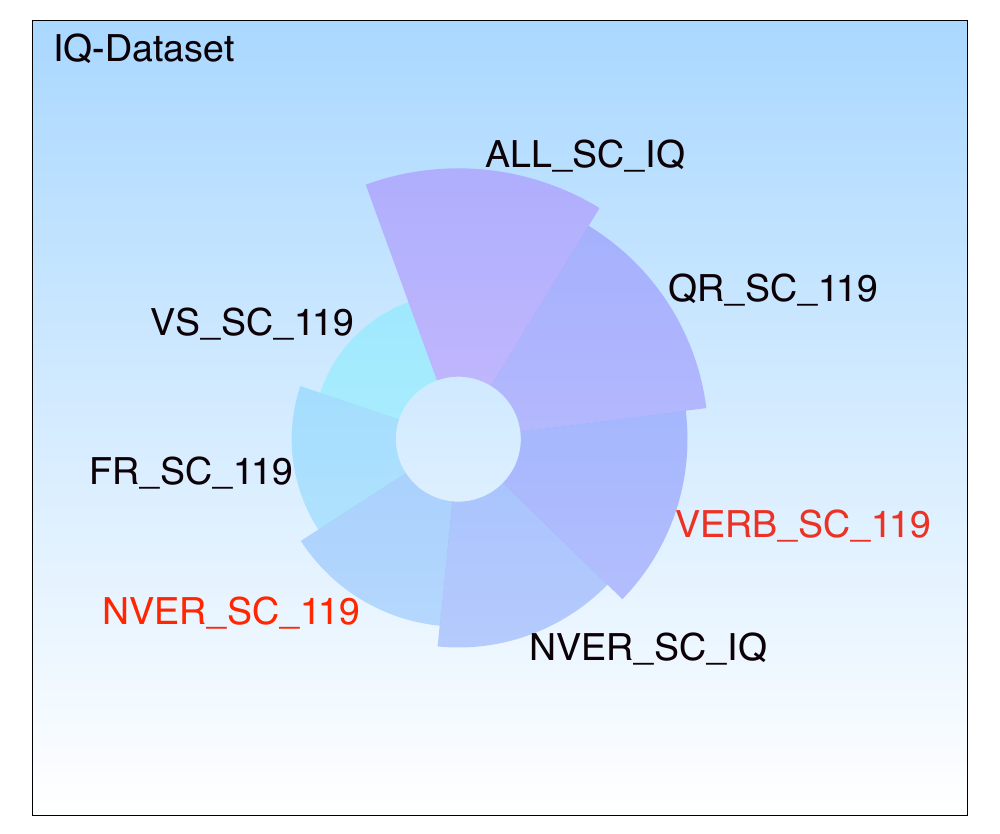}
        }
        \vspace{-0.2cm}
        \caption{Case Study. Traceability comparison of feature subsets selected by {\model} and {\extendedmodel}.} 
        \label{case_study}
\end{figure}

We conduct this experiment to examine the traceability of {\model} and {\extendedmodel}. 
We rank the top 7 significant features for prediction in the feature subset selected by {\model} and {\extendedmodel} using the IQ-Dataset.
Figure~\ref{case_study} (a-b) show the top 7 significant features among the selected features of {\model} and {\extendedmodel} respectively. 
The predictive label is the summation of NVER\_SC\_119 and VERB\_SC\_119, which is the summed standardized score from 1 to 19 for the Nonverbal Intelligence scale and the summed standardized score from 1 to 19 for the Verbal Intelligence scale respectively. From Figure~\ref{case_study} we observe that both of {\model} and {\extendedmodel} successfully selected these two features from 38 original features. 
A potential reason for this observation is that both of {\model} and {\extendedmodel} are capable of capturing the complicated interactions among features as well as the causality between features and predictive label. 
Thus, this experiment successfully demonstrates the traceability of the feature subset selected by {\model} and {\extendedmodel}, highlighting the effectiveness from different a perspective.

\vspace{0.3cm}
\section{Related Works}
\vspace{0.4cm}
\noindent\textbf{Feature selection} (FS) screens informative features from raw data, improving model efficiency and performance.
Existing FS methods are grouped into three types: 
1) Filter methods evaluate feature importance using statistical metrics (e.g., correlation and mutual information)~\cite{kbest,mrmr}.
However, such methods generally ignore the dependency and interactions among features.
2) Wrapper methods formulate FS as a search problem, training a machine learning (ML) model as evaluation metric and selecting the subset with the best performance~\cite{liu2021efficient,gfs}. However, such methods require excessive model training, resulting in high computational costs. 
3) Embedded methods embed FS into model training to assess the importance of features~\cite{lasso,sugumaran2007feature,lassonet,rfe}. However, such methods heavily rely on model structure. 
Recently, generative AI have attracted substantial attention~\cite{gains,ying2024feature,grfg,2022traceable,2023traceable,ying2025distribution,huang2025collaborative,gong2025neuro,ref_arxiv,urbanplanning_rui,catf_rui}.
For instance,~\cite{gains} proposes a framework that employs a sequential encoder, an accuracy evaluator, a sequential decoder, and a gradient ascent optimizer, embedding discrete FS knowledge into a continuous embedding space and searching the space for the optimal feature subset with gradient-ascent method.
However, this method is limited by: 
1) incapable of encoding the permutation invariance into the embedding space; 
2) the reliance on the convexity assumptions of the embedding space.
\noindent\textbf{Federated learning} (FL) enables multiple clients to train a shared model without sharing the raw data, addressing privacy and security concerns in distributed settings. 
Existing FL algorithms rely on local model updates that are aggregated by a central server across communication rounds, focus on parameter-level aggregation. 
FedAvg~\cite{fedavg} enables each client to perform multiple steps of stochastic gradient descent (SGD) on its local dataset before transmitting model parameters to the server. 
The server then computes a weighted average of the client models, assigning each client a weight based on the proportion of its local sample size to the total number of samples across all participating clients.
FedNTD~\cite{fedntd} incorporates a not-true distillation mechanism, where each client optimizes its local model with respect to both the ground-truth labels and the non-true prediction distribution provided by the global model. 
The additional not-true labels enable more knowledge transfer across clients and reduces the negative impact of data heterogeneity. 
FedProx~\cite{fedprox} utilizes a proximal regularization term to the local optimization objective, constraining client updates to remain close to the global model parameters.
MOON~\cite{moon} introduces a model-contrastive learning objective that regulates local training.
Specifically, each client is encouraged to align the representation learned by its local model with the representation learned by current global model, while simultaneously discouraging similarity to the representation of its previous local model.
These methods requires frequent transmission of large parameter sets and high communication overhead and ignores diverse FS knowledge within local datasets.
To fundamentally overcomes these limitations, we integrate a permutation-invariant embedding mechanism with a policy-guided search strategy in {\model} for the centralized scenarios and further extend {\model} to the federated settings.
The centralized version {\model} integrates permutation-invariant embedding to mitigate the sensitivity to feature ordering and a policy-guided search strategy to effectively explore this space.
The federated version {\extendedmodel} further introduce a privacy-preserving knowledge fusion strategy combined with a sample-aware weighted aggregation strategy to integrate FS knowledge across clients while mitigating the bias introduced by heterogeneous distributions of client data.

\vspace{-0.2cm}
\section{Conclusions}
We propose a generative automated feature selection framework in both centralized and federated settings.
The centralized model integrates permutation-invariant embeddings and policy-guided search to overcome permutation bias and convexity assumptions.
In detail, we first train a permutation-invariant feature subset embedding module using feature selection records by optimizing the reconstruction loss. The encoder maps feature subsets to continuous embeddings, while the decoder reconstructs the subsets from these embeddings.
Moreover, considering the high computational cost of the pairwise interaction calculations, we leverage a set of inducing points as intermediate representations. 
These inducing points are expected to extract information about the input feature subset, contributing to capture global feature pattern and facilitate more efficient attention calculations by reducing the need for full pairwise attention. 
As the model converges, we employ a policy-guided RL agent to explore the learned space for superior feature subsets, overcoming the reliance on convex assumptions and mitigating the risk of converging to local optima. 
The federated model extends this framework with privacy-preserving knowledge fusion and sample-aware weighted aggregation, enabling effective knowledge transfer across heterogeneous clients while reducing computation overhead.
Finally, extensive experiments and comprehensive analyses demonstrate the effectiveness of the proposed models.


\balance

\bibliographystyle{IEEEtran}
\bibliography{IEEEabrv,IEEEexample}

\end{document}